\documentclass[conference]{IEEEtran}

\usepackage{amsfonts}
\usepackage{amssymb}
\usepackage{booktabs}
\usepackage{graphicx}
\usepackage[bookmarks=true]{hyperref}
\usepackage{multicol}
\usepackage[numbers]{natbib}
\usepackage{caption,subcaption}
\usepackage{siunitx}
\usepackage{tikz,graphics,float,epsf}
\usepackage{times}
\usepackage{xcolor}
\usepackage{bm}
\usepackage{bbm}
\usepackage{amsmath}
\usetikzlibrary{calc}
\usepackage{xspace}

\definecolor{es-blue}{rgb}{0,0.4,0.8}

\makeatletter
\newcommand{\printfnsymbol}[1]{%
  \textsuperscript{\@fnsymbol{#1}}%
}
\makeatother
\usepackage[symbol]{footmisc}

\newcommand{\vxcmd}{\textbf{v}_x^{\textrm{\footnotesize {cmd}}}}
\newcommand{\vycmd}{\textbf{v}_y^{\textrm{\footnotesize {cmd}}}}
\newcommand{\wzcmd}{\boldsymbol{\omega}_z^{\textrm{\footnotesize {cmd}}}}
\newcommand{\at}{\textbf{a}_{t}}
\newcommand{\vt}{\textbf{v}^{\footnotesize{\textrm{cmd}}}_t}
\newcommand{\dt}{\textbf{d}_t}
\newcommand{\xt}{\textbf{x}_t}
\newcommand{\zt}{\textbf{z}_t}
\newcommand{\zht}{\hat{\textbf{z}}_t}
\newcommand{\Rnum}{\mathbb{R}}
\newcommand{\rwz}{r_{\omega^{\footnotesize{\textrm{cmd}}}_z}}
\newcommand{\rvx}{r_{v^{\footnotesize{\textrm{cmd}}}_x}}

\newcommand{\tikzmark}[1]{\tikz[overlay,remember picture] \node (#1) {};}
\newcommand{\DrawBox}[3][]{%
    \tikz[overlay,remember picture]{
    \draw[black,#1]
      ($(#2)+(-0.5em,2.0ex)$) rectangle
      ($(#3)+(0.75em,-0.75ex)$);}
}

\begin{document}

\title{Rapid Locomotion via Reinforcement Learning}

\author{\authorblockN{Gabriel B. Margolis\printfnsymbol{1}$^1$,
Ge Yang\printfnsymbol{1}$^1$ $^2$,
Kartik Paigwar$^1$, 
Tao Chen$^1$, and
Pulkit Agrawal$^1$ $^2$}
\authorblockA{$^1$MIT Improbable AI Lab \quad $^2$NSF AI Institute for Artificial Intelligence and Fundamental Interactions  \\ Massachusetts Institute of Technology,
Cambridge, MA 02139 
}
}

\maketitle

\begin{abstract}

Agile maneuvers such as sprinting and high-speed turning in the wild are challenging for legged robots. 
We present an end-to-end learned controller that achieves record agility for the MIT Mini Cheetah, sustaining speeds up to \SI{3.9}{\meter/\second}. This system runs and turns fast on natural terrains like grass, ice, and gravel and responds robustly to disturbances. Our controller is a neural network trained in simulation via reinforcement learning and transferred to the real world. The two key components are (i) an adaptive curriculum on velocity commands and (ii) an online system identification strategy for sim-to-real transfer leveraged from prior work. Videos of the robot's behaviors are available at \url{https://agility.csail.mit.edu/}.

\end{abstract}

\IEEEpeerreviewmaketitle

\footnotetext[1]{\fontsize{8}{8} Equal contribution. Authors are also affiliated with Computer Science and Artificial Laboratory (CSAIL), the Laboratory for Information and Decision Systems (LIDS), and the MIT-IBM Watson AI Lab at MIT. Correspondence to {\href{mailto:gmargo@csail.mit.edu,geyang@csail.mit.edu}{\texttt{\{gmargo, geyang\}@csail.mit.edu}}}}

\section{Introduction}
\label{sec:intro}

Running fast on natural terrain is challenging. 
Different terrains exhibit different characteristics, ranging from variable friction and softness to sloped and uneven geometry.
As a robot attempts to move at faster speeds, the impact of terrain variation on controller performance increases~\cite{bosworth2016robot, fahmi2020stance}. Some physical considerations only begin to influence the robot's dynamics at high speeds, including the enforcement of
actuator limits~\cite{chignoli2021humanoid, dai2014whole, herzog2016structured}, the regulation of large contact forces~\cite{kim2019highly}, and body control during flight phases~\cite{dai2014whole, kim2019highly}.
One possibility is to resolve these issues by making targeted improvements to the hand-designed models used in model-based control. Impressive progress has been made in this direction~\cite{bledt2018cheetah, bosworth2016robot, chignoli2021humanoid,dai2014whole, ding2019realtime, fahmi2020stance, herzog2016structured, kim2019highly}. However, in model-based control, the robot's behavior and robustness are dependent on the creativity and investment of the human designer, who must invent simplified reduced-order models that
allow the robot to infer the appropriate actions 
under the constraint of real-time computation.

\begin{figure}[t!]
    \centering
    \includegraphics[width=0.33\linewidth]{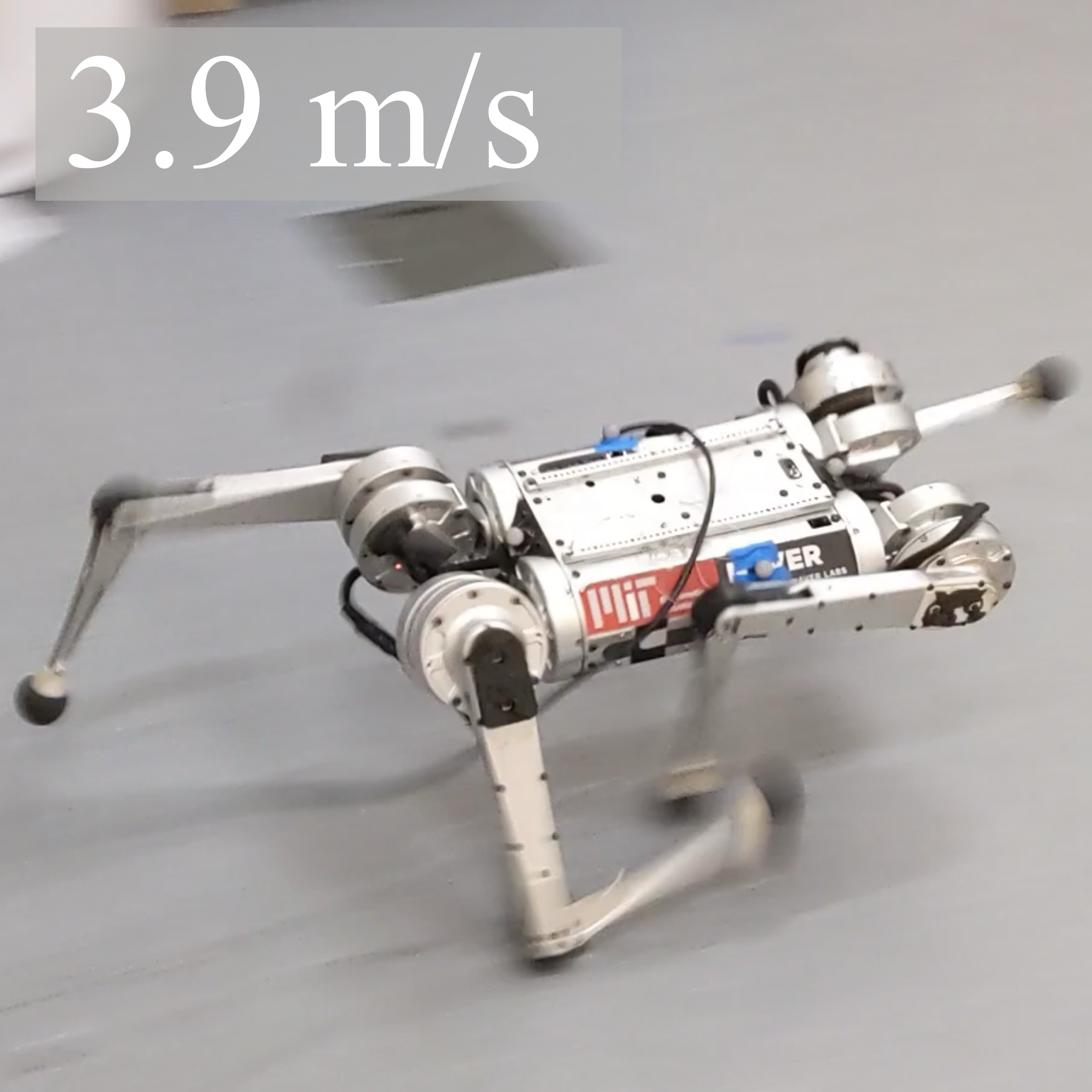}\hfill
    \includegraphics[width=0.33\linewidth]{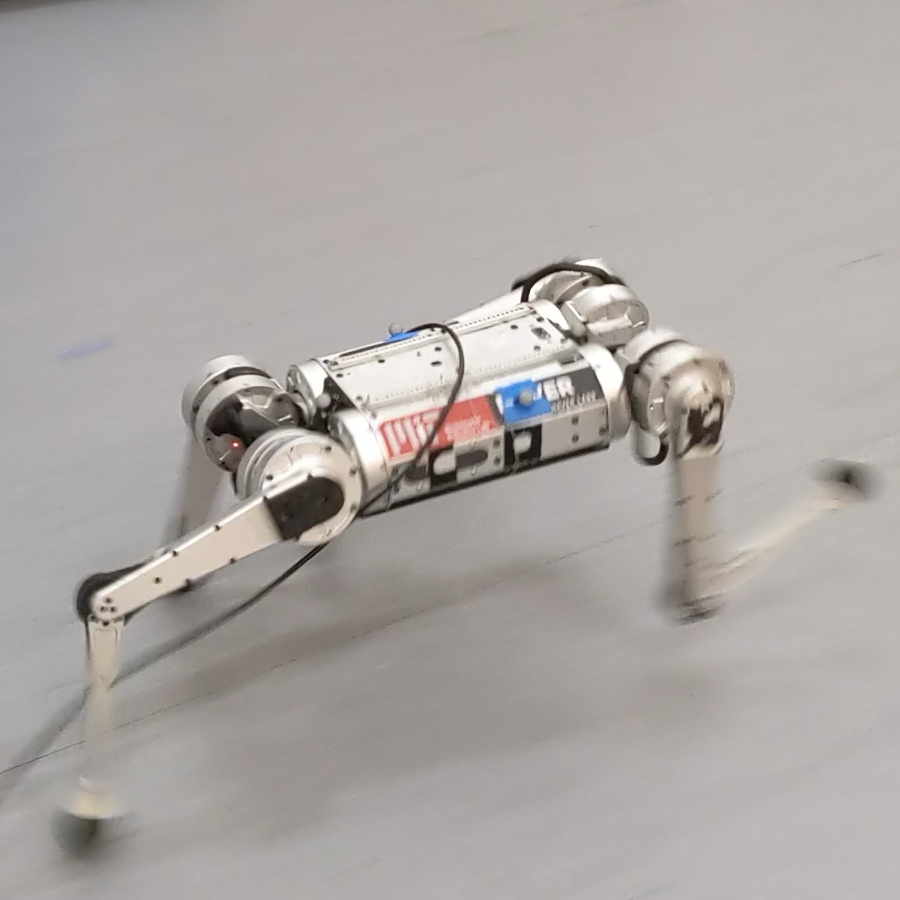}\hfill
    \includegraphics[width=0.33\linewidth]{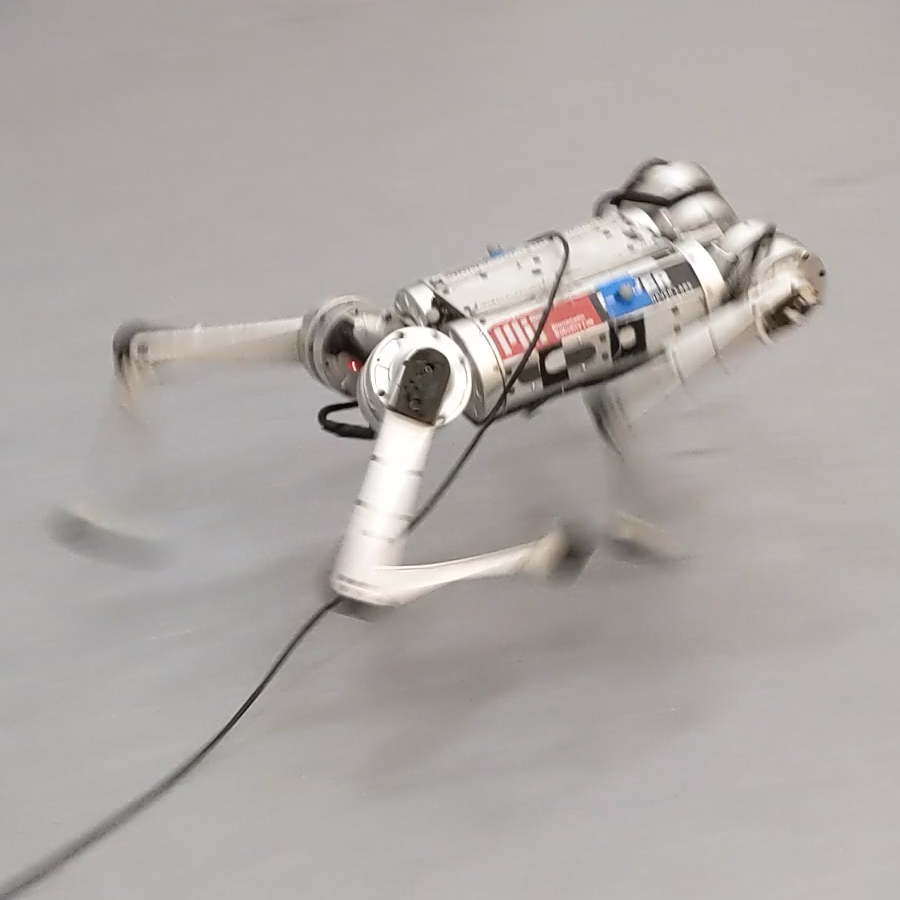}\\ \vspace{0.1cm}
    \includegraphics[width=0.33\linewidth]{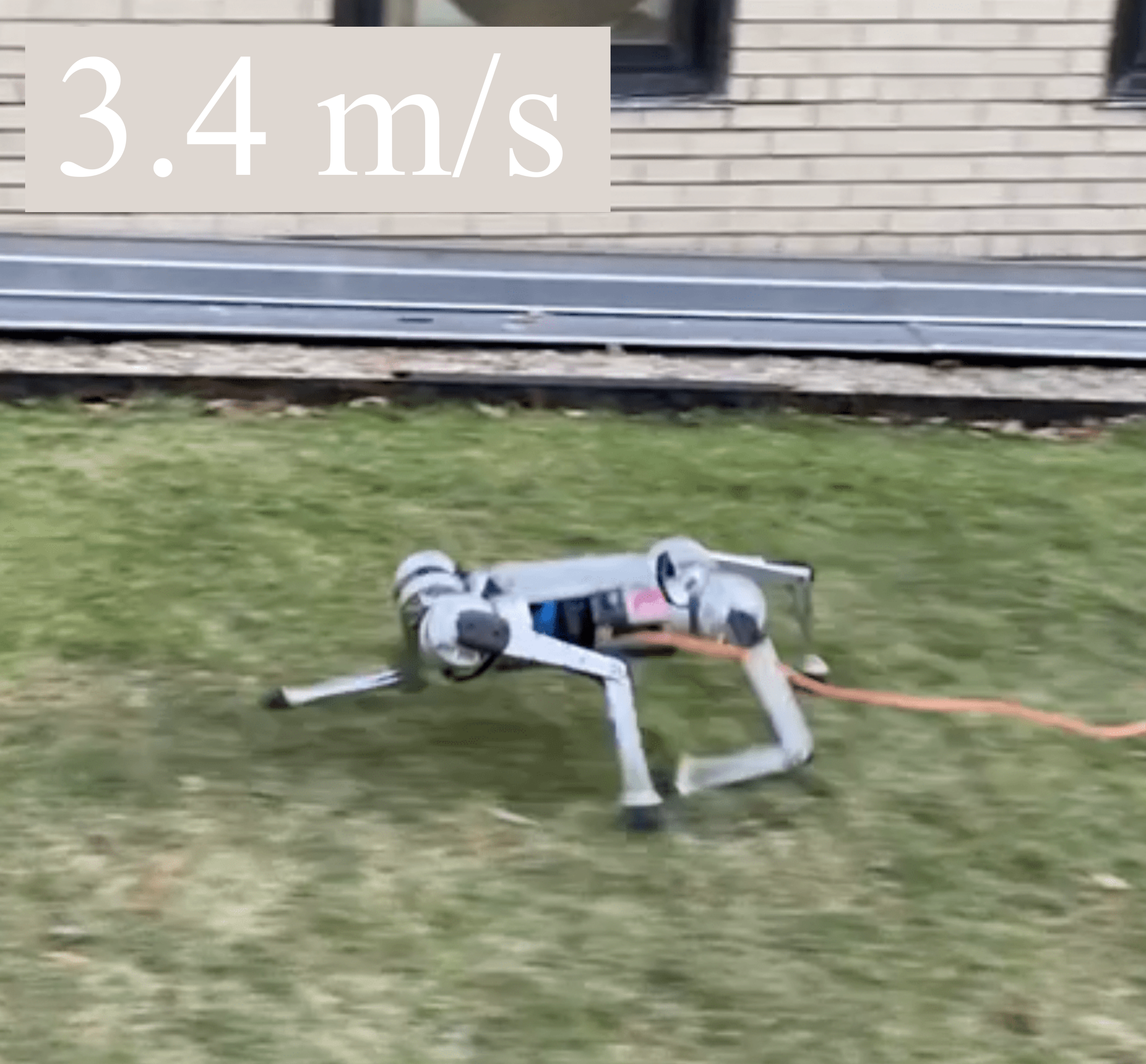}\hfill
    \includegraphics[width=0.33\linewidth]{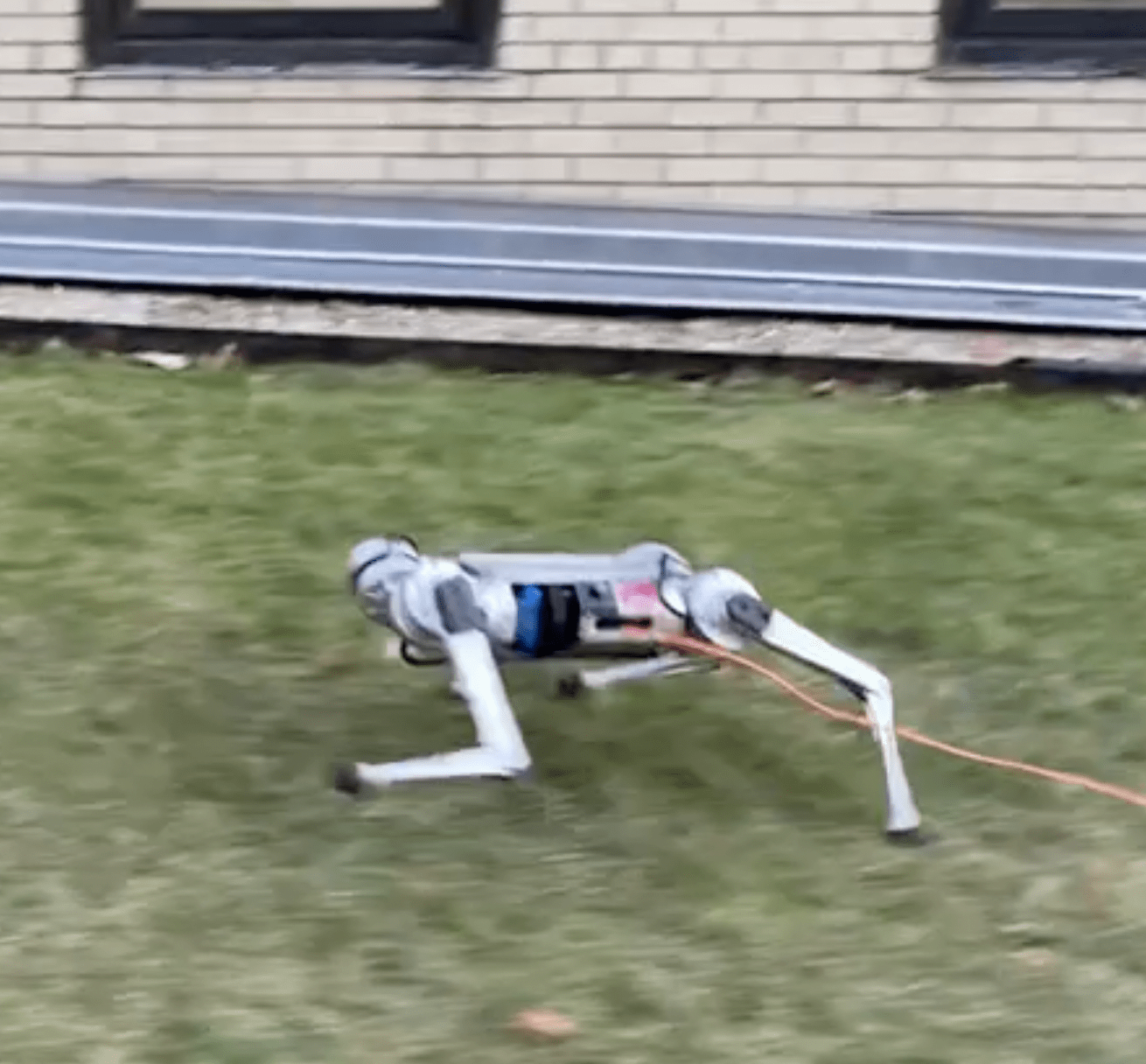}\hfill
    \includegraphics[width=0.33\linewidth]{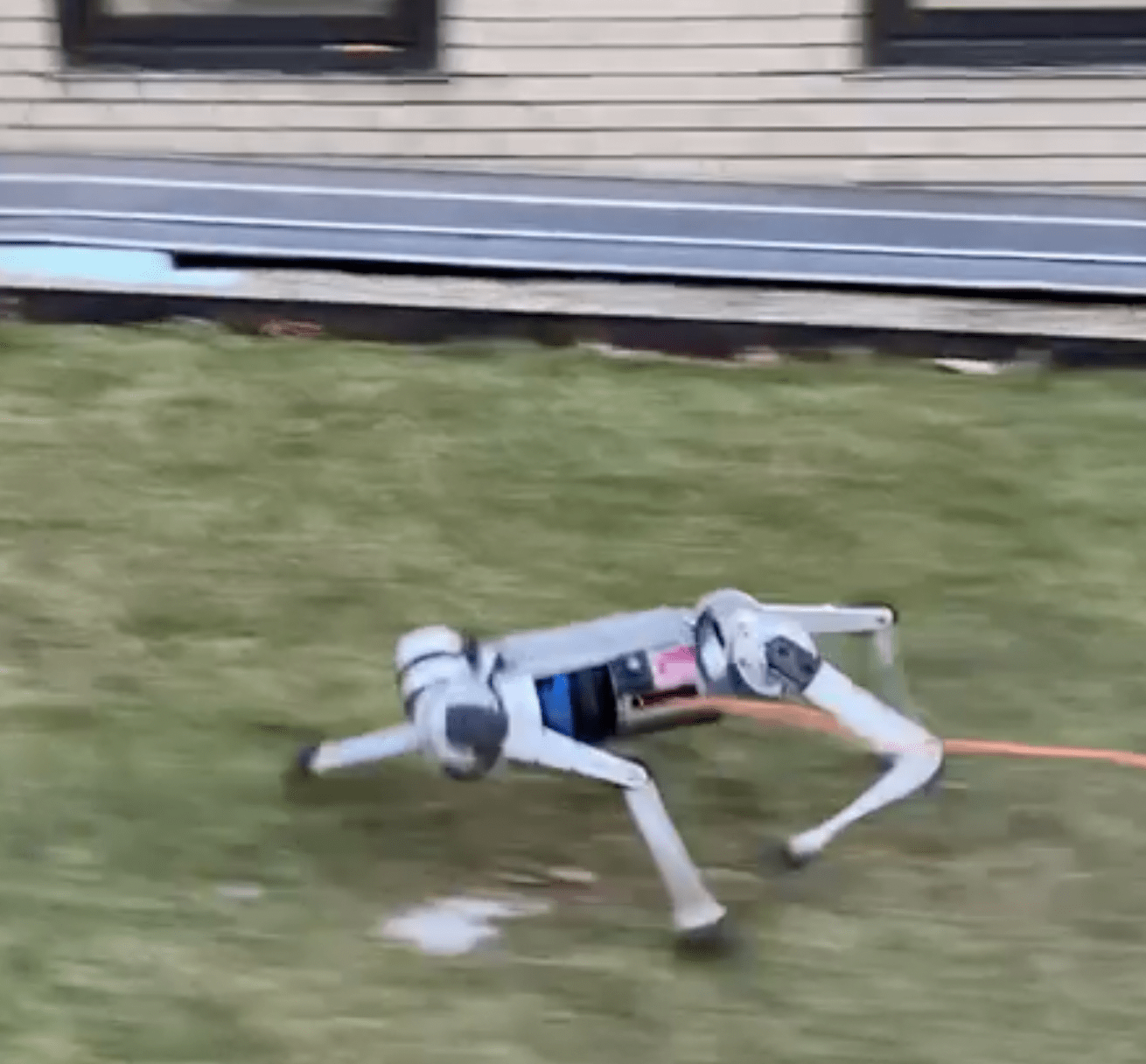}\\ \vspace{0.1cm}
    \includegraphics[width=0.33\linewidth]{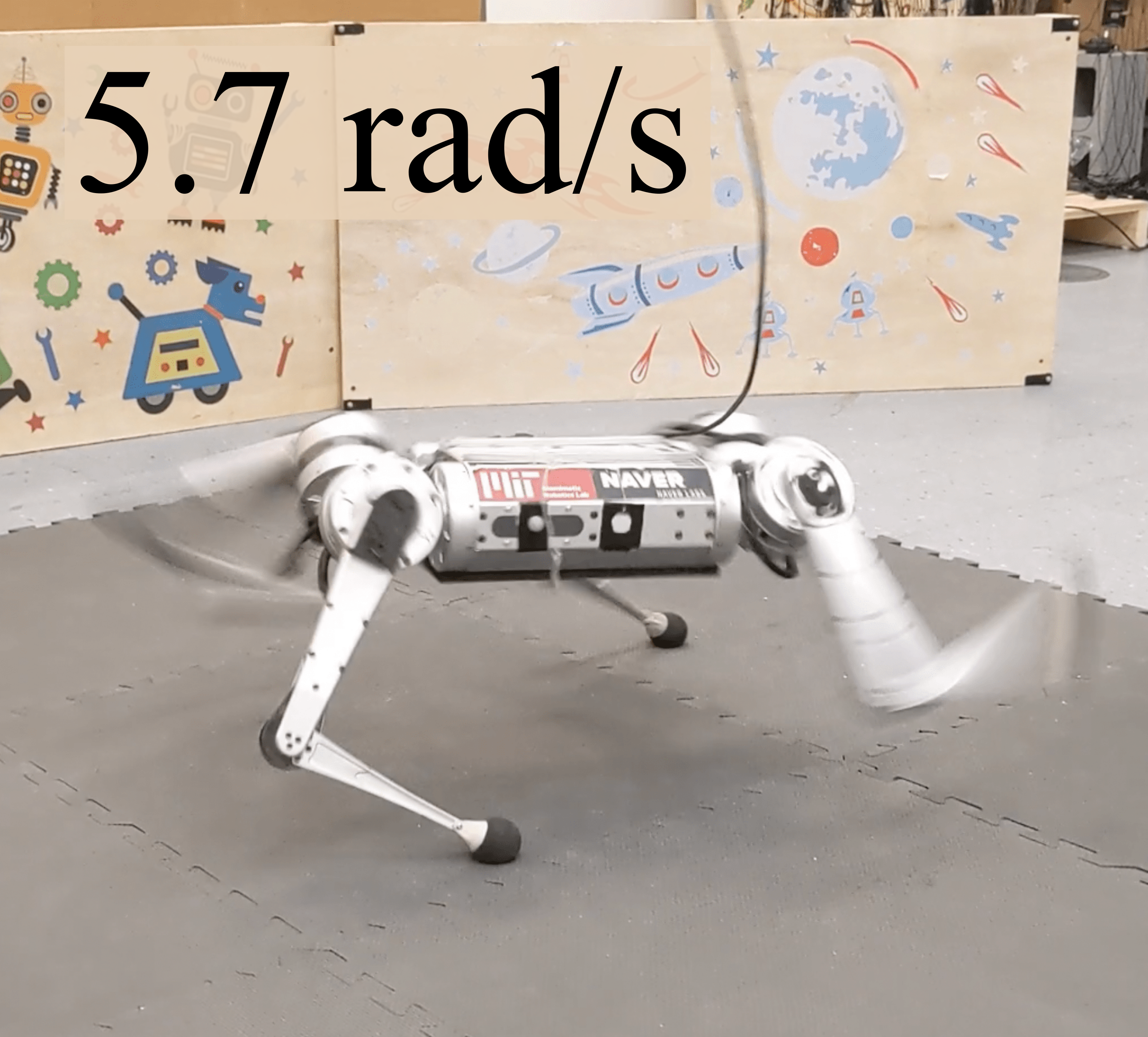}\hfill
    \includegraphics[width=0.33\linewidth]{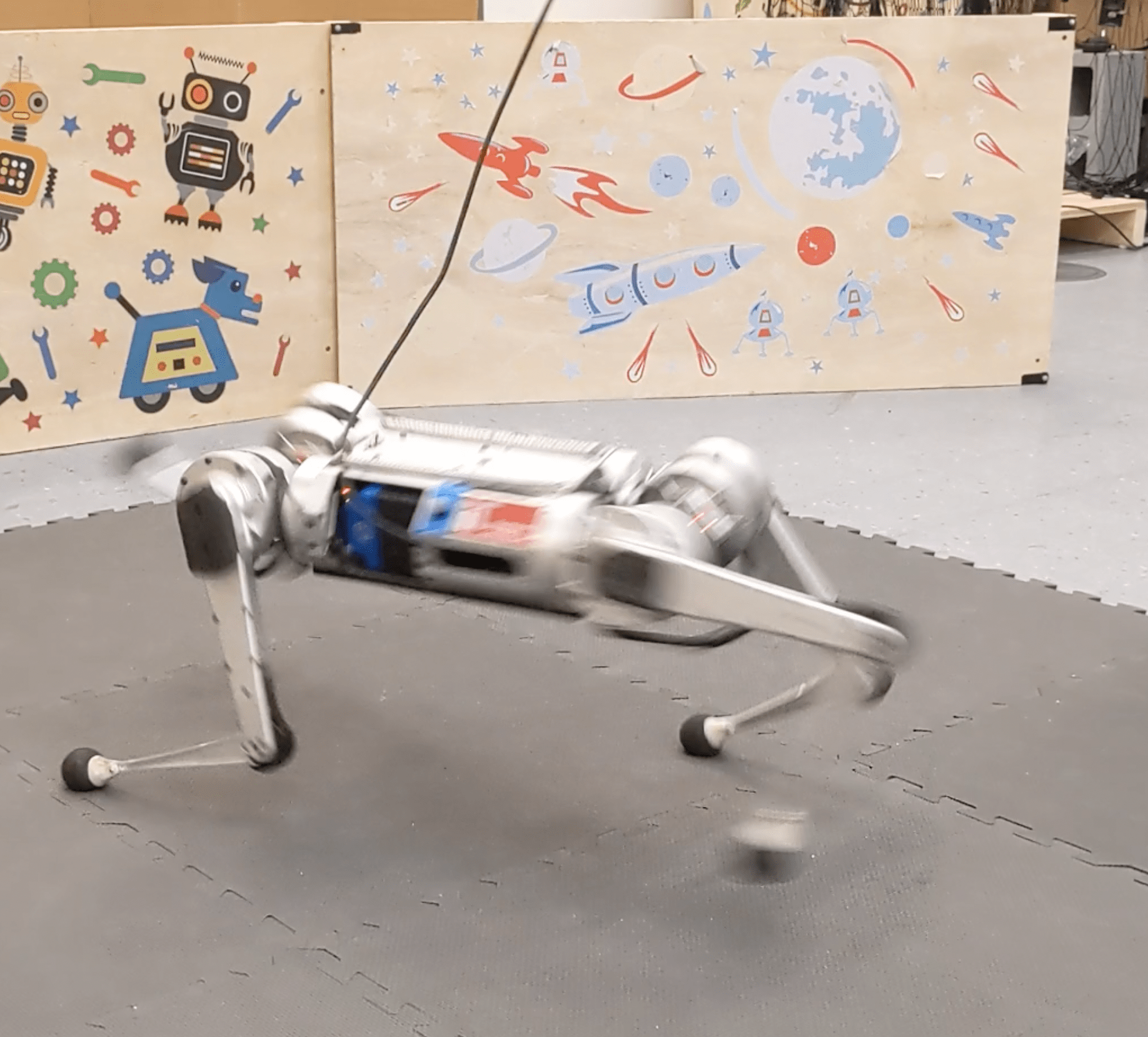}\hfill
    \includegraphics[width=0.33\linewidth]{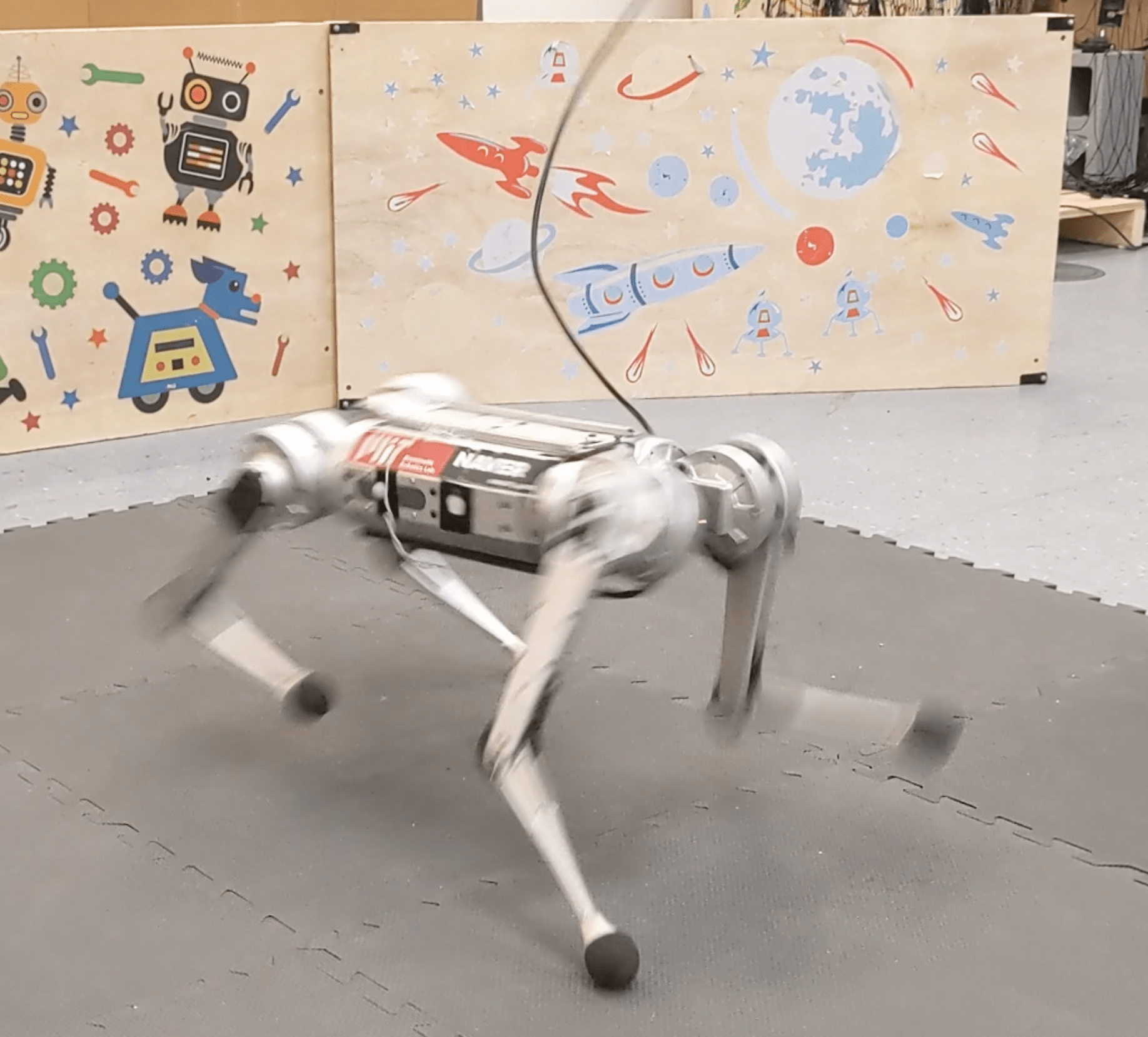}\\ \vspace{0.1cm}
    \includegraphics[width=0.33\linewidth]{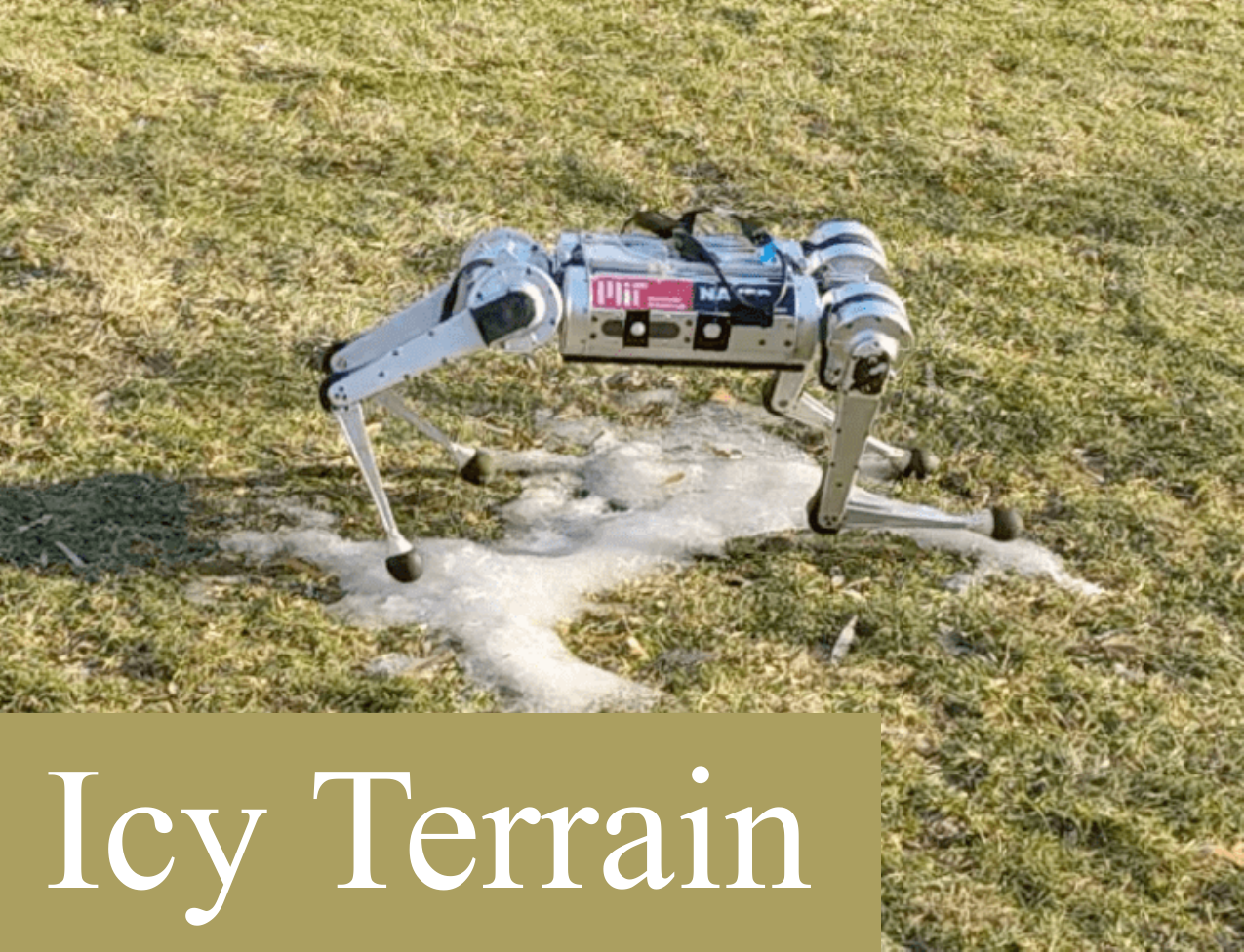}\hfill
    \includegraphics[width=0.33\linewidth]{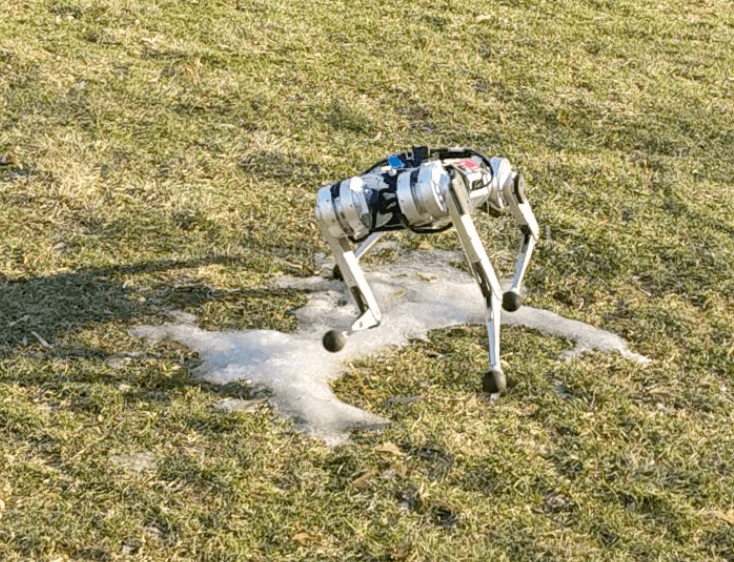}\hfill
    \includegraphics[width=0.33\linewidth]{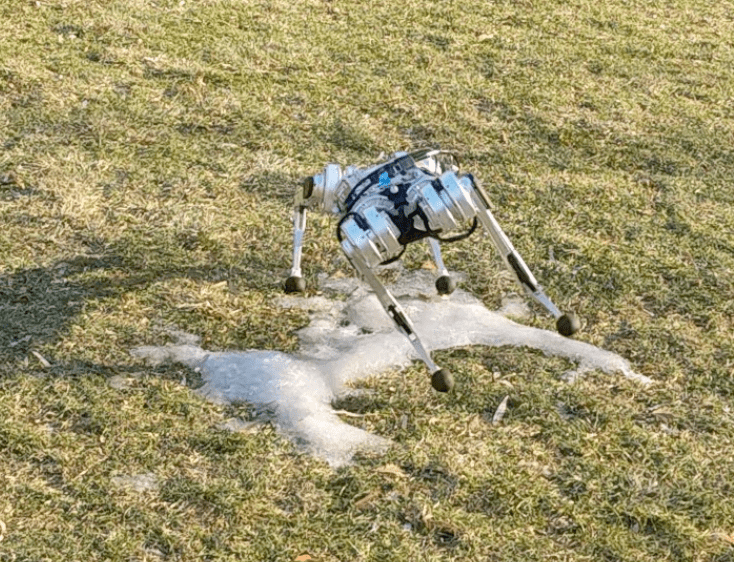}\\
    \caption{
    An end-to-end learned controller enables the MIT Mini Cheetah to execute: (a) fast sprinting at \SI{3.9}{\meter/\second} (top); (b) a rough terrain \(10\)-meter sprint at \SI{3.4}{\meter/\second}; (c) high-speed spinning indoors; and (d) robust spinning on an icy patch. All behaviors are realized by a single neural network that is trained in simulation and deployed zero-shot in the real world.
    }\label{fig:teaser}
    \vspace{-0.6cm}
\end{figure}

How can we perform real-time control in complex environments where efficient reduced-order models may not exist or are currently unknown? 
One possibility is to optimize the robot's actions with respect to a full physics model.
The problem is that trajectory optimization with a full model is not possible in real-time for a complex task such as fast running on natural terrains. An alternative is to amortize the cost of trajectory optimization by learning a direct mapping from sensory observations to actions (a \textit{policy}) using high-reward trajectories sampled from the full model. Reinforcement learning (RL) provides a way to learn such a policy. 
In this approach, the human designs a set of training environments
and reward functions to specify a set of tasks. RL algorithms automatically discover the policy that maximizes reward across these environments and tasks.
Because the RL framework does not require a human engineer to design accurate and efficient reduced-order models, it is less reliant on human effort. Consequently, RL offers a scalable controller synthesis scheme for complex tasks in challenging environments. Recent works have successfully employed RL to learn locomotion controllers~\cite{kumar2021rapid, lee2020learning, margolis2021learning, miki2022learning, rudin2021learning, siekmann2021blind}. 

Our goal is to construct a system that can traverse terrains at a large range of linear and angular velocities. This corresponds to a multi-task RL setup where running with each combination of linear and angular velocity constitutes a separate task. 
Akin to prior work, we found that when the robot is trained to walk with a narrow range of commanded velocities, a multi-task policy can be successfully learned~\cite{kumar2021rapid,lee2020learning}. However, increasing the range of commanded velocities to include high speeds results in training failure. This issue is reminiscent of difficulty in learning multi-task policies via RL on a broad set of tasks~\cite{hessel2019multi}. 
To understand the reason for failure, consider the naive approach of training a multi-task RL policy by uniformly sampling from all tasks. If most of the tasks are challenging or infeasible, the agent will not gather significant reward, and the training will fail. This is the case in high-speed locomotion: learning to run at rapid velocities from scratch is difficult because physical considerations such as centrifugal force constrain the combinations of linear and angular speed that are realizable. 

Training can be made easier by initially providing simple tasks to the agent and then slowly increasing their complexity using a curriculum~\cite{bengio2009curriculum}. Curriculum learning has been leveraged for training robotic systems in the past~\cite{li2020towards, matiisen2019teacher, rudin2021learning, xie2020allsteps}. Manual curriculum design can fail when
the difficulty or feasibility of tasks is not known in advance. For omnidirectional running, manual curriculum design involves finding feasible linear and angular velocity combinations that satisfy physical constraints and ranking velocity commands based on their difficulty. Task difficulty is a function of both the system dynamics and the optimization algorithm, making manual curriculum design tedious and problem-dependent. Instead, we implement an automatic curriculum strategy that expands the set of tasks while respecting the physical constraints of locomotion. The proposed strategy yields significant performance improvements in learning omnidirectional high-speed locomotion.
  
When deployed in the real world on flat ground, our learned policy sustained a top speed of \textbf{\SI{3.9}{\meter/\second}}, the highest reported speed for this robot (first row of Figure~\ref{fig:teaser}). On uneven outdoor terrain covered with grass, our robot achieved an average speed of \SI{3.4}{\meter/\second} for a \SI{10}{\meter} dash (second row of Figure~\ref{fig:teaser}). The same policy can spin the robot at  \textbf{\SI{5.7}{\radian/\second}} on flat ground and also enables the robot to spin on the more challenging icy terrain (bottom row of Figure~\ref{fig:teaser}). We observed additional emergent behaviors during our experiments, including recovery from tripping and compensation for a malfunctioning motor. These results are reported qualitatively, and corresponding videos highlight the diversity of responses that emerge from end-to-end learning.  

Our policy uses a minimal sensing suite, consisting only of gyroscope and joint encoders, and is therefore suitable for any typical robot quadruped, including relatively inexpensive commercially available robots.
Overall, our system performs rapid locomotion both indoors and outdoors and successfully negotiates
challenging terrains and disturbances.
Our work fills a gap in the literature. 
We show that reinforcement learning can be used to learn locomotion controllers that simultaneously achieve linear and angular high-speed behaviors and operate on diverse natural terrains.

\section{Experimental Setup}

\noindent \textbf{Hardware:} We use the MIT Mini Cheetah \cite{katz2019platform} as our experimental platform. The robot stands \SI{30}{\cm} tall and weighs \SI{9}{\kilogram}. It is equipped with \(12\) quasi-direct-drive actuators capable of maximum output torque of \SI{17}{\newton\meter}. The robot's sensor suite consists of joint position encoders and an inertial measurement unit (IMU). Our neural network controller runs at \SI{50}{\hertz} on an onboard NVIDIA Jetson TX2 NX computer.

\noindent \textbf{Simulation:} We use the IsaacGym simulator~\cite{makoviychuk2021isaac} 
and code adapted from the open-source repository
in \cite{rudin2021learning}. We collect \(400\) million simulated timesteps using \(4000\) parallel agents for policy training. This is roughly equivalent to \(92\) real-time days, which we can simulate in under three hours of wall-clock time using a single NVIDIA RTX 3090 GPU.

\section{Method}
\label{sec:method} 

\begin{figure}
    \centering
    \includegraphics[width=\linewidth]{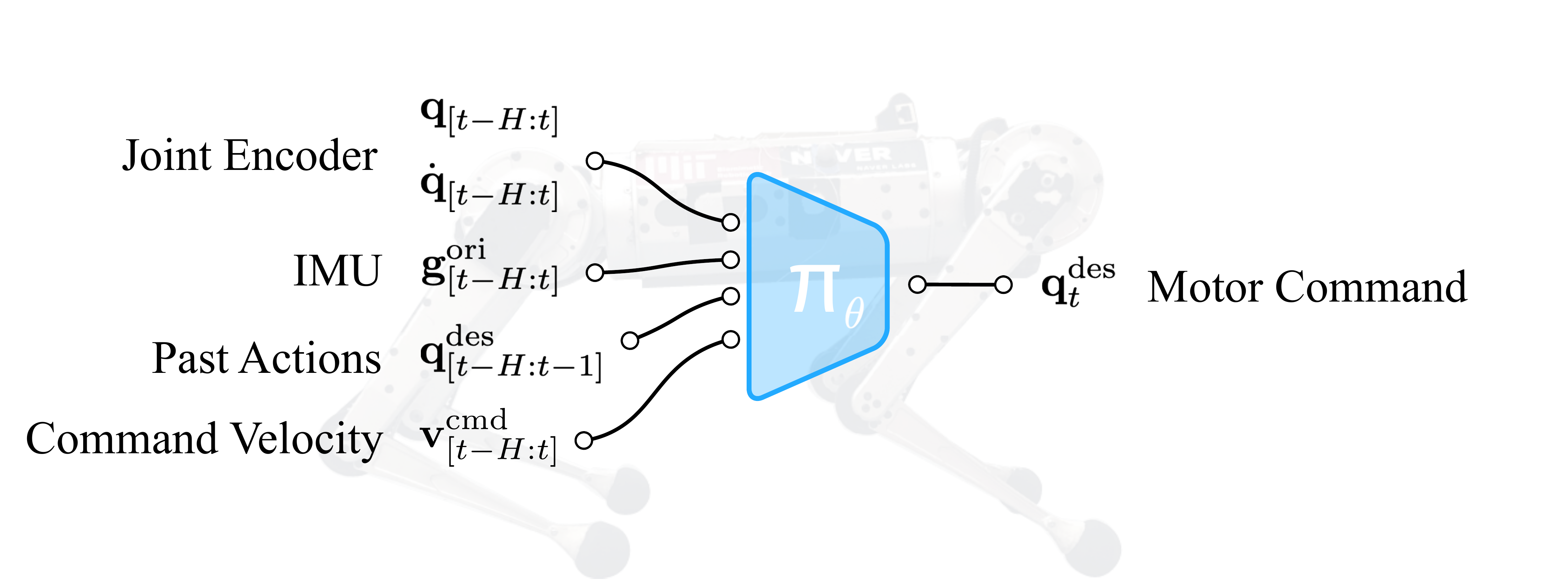}
    \caption{Our controller is a learned mapping from sensory inputs to desired joint positions. We parameterize it as 5-layer neural network $\pi_{\theta}$ with parameters $\theta$ optimized in simulation.}
    \label{fig:arch}
\end{figure}

Our goal is to learn a policy $\pi_{\theta}(.)$ with parameters $\theta$ that takes as input sensory data and velocity commands and gives as output joint position commands (see Figure~\ref{fig:arch}), which are converted into joint torques by a PD controller. The command ($\vt$) includes the longitudinal ($\vxcmd$) and lateral ($\vycmd$) linear velocities and the yaw rate ($\wzcmd$).

\subsection{Control Architecture}
\label{sec:control-architecture}

\noindent \textbf{Observation Space:} The robot's sensors provide joint angles ${\textbf{q}_t \in\Rnum^{12}}$ and joint velocities ${\dot{\textbf{q}}_t\in\Rnum^{12}}$, measured using motor encoders, and ${\textbf{g}^{\footnotesize{\textrm{ori}}}_t\in\Rnum^{3}}$, which denotes the orientation of the gravity vector in the robot's body frame and is measured using the IMU. 
As detailed in Section \ref{sec:method-policy-architecture}, the policy $\pi_{\theta}(\cdot)$ takes as input a history of previous observations and actions denoted by $\textbf{o}_{t-H:t}$ where ${\textbf{o}_t = [\textbf{q}_t, \dot{\textbf{q}}_t, \textbf{g}^{\footnotesize{\textrm{ori}}}_t, \textbf{a}_{t-1}]}$. Because we are learning a command-conditioned policy, the input to the policy is ${\textbf{x}_{t-H:t}}$ where ${\xt = \textbf{o}_t \bigoplus  \vt}$. During deployment, the body velocity command $\vt$ is specified by a human operator via remote control.

\noindent \textbf {Action Space:} The action, $\at \in \mathbb{R}^{12}$, assigns joint position commands for a PD controller. The proportional gain is \(20\), and the derivative gain is \(0.5\). We chose these low gains to promote smooth motions and did not tune them during our experiments. 

\noindent \textbf{Reward Function} closely follows~\cite{rudin2021learning} with task reward terms for linear and angular velocity tracking, as well as a set of auxiliary terms for stability (velocity penalties on body roll, pitch, height), smoothness (joint torque and acceleration penalties, action change penalty, footswing duration bonus), and safety (penalty on self-collision, penalty on joint limit violations). We found that the robot tends to sink its body at high speeds and lean into its heading. This motivated us to introduce penalties on the robot's body height and orientation.
The details of the reward function are in Table~\ref{tbl:rewards} (Appendix).

\subsection{Teacher-Student Training
}
\label{sec:method_opt}

We train a locomotion policy in simulation and transfer it to the real world without fine-tuning. Because the real-world terrain and some of the robot's parameters are not precisely known, it is common practice to train $\pi_{\theta}(.)$ by randomizing simulation parameters denoted here as $\dt$. We randomize the body mass, the center of mass, motor strength, ground friction, and ground restitution in the ranges reported in Table~\ref{tab:domain_rand}.

\begin{table}[t!]
\centering\small
\begin{tabular}{lccl}
\toprule
            Term           &  Min    &  Max &       Unit   \\
\midrule
Ground Friction            & 0.05    & 4.00 &  \;\;   -    \\
Ground Restitution         & 0.00    & 1.00 &  \;\;   -    \\
Payload Mass               & -1.0    &  3.0 &  \;  \SI{}{\kilogram}  \\
Body Center of Mass        & -0.10   & 0.10 &  \;   \SI{}{\meter}  \\
Motor Strength             &  90     &  110 &  \;  \(\%\)  \\ \hline
Forward Velocity Command ($\vxcmd$)   &  var.   & var. & \;\;\SI{}{\meter/\second}  \\
Lateral Velocity Command ($\vycmd$)   &  var.   & var. & \;\;\SI{}{\meter/\second}  \\
Angular Velocity Command ($\wzcmd$)  &  var.   & var. &   \;\;\SI{}{\radian/\second}  \\
\bottomrule
\end{tabular}
\caption{The first set of rows report the ranges of the domain parameters we randomize. The policy is tasked to follow a range of velocity commands that are generated via curriculum strategy described in Section~\ref{sec:method_curriculum}.}
\label{tab:domain_rand}
\end{table}

One possibility is for the policy to learn a single behavior that works across all the randomized parameters. This learning procedure is commonly referred to as \textit{domain randomization}~\cite{tan2018simtoreal, tobin2017domain}. Let the resulting policy be $\pi_{DR}(\xt)$. While $\pi_{DR}(\xt)$ can cross the sim-to-real gap~\cite{tan2018simtoreal, tobin2017domain}, the learned behavior is conservative~\cite{tan2018simtoreal,xie2021dynamics} because there is no mechanism for the policy to adapt to different domain parameters. For instance, from the same starting state, it makes sense to run on ice in a manner different from running on grass. However, $\pi_{DR}(\xt)$ has no mechanism for doing so. 

To prevent the policy from being conservative, one approach is to include the domain parameters $\dt$ as part of the policy input~\cite{chen2020learning}. The policy ${\pi_T(\xt, \dt)}$, commonly referred to as a
\textit{teacher policy}, is trained using an RL algorithm to maximize the expected sum of rewards. Direct knowledge of the domain parameters provides $\pi_T(\xt, \dt)$ with the ability to adapt to different domains. However, this policy cannot be deployed on a real robot since $\dt$ cannot be directly measured using onboard sensors. To overcome this limitation, one can deploy a \textit{student policy}, $\pi_S(\xt, \textbf{x}_{[t-h:t-1]})$ that is trained to mimic the teacher's action via behavior cloning~\cite{ross2011reduction}. The main idea is that accurately matching the teacher's actions forces the student to implicitly infer domain parameters ($\dt$) from a state history of $h$ time steps, \(\textbf{x}_{[t-h:t-1]}\). Therefore, the student policy is said to perform online system identification. 

Teacher-student training enables the agent to specialize its behavior to the current dynamics $\dt$, instead of learning a single behavior that works across different $\dt$. This so-called implicit system identification approach has been previously developed in a number of works involving object re-orientation with a multi-finger hand~\cite{chen2021system}, self-driving cars~\cite{chen2020learning} and locomotion~\cite{kumar2021rapid,lee2020learning,margolis2021learning,miki2022learning}. Like work applying student-teacher learning to blind walking~\cite{kumar2021rapid, lee2020learning}, our teacher policy observes $\dt$, the dynamic properties of the robot and terrain. The student learns to infer them from \(\textbf{x}_{[t-h:t-1]}\), the history of joint angles, and IMU readings.

\subsection{Policy Optimization}
\label{sec:method-policy-architecture}

\subsubsection{Teacher Policy}
We construct the teacher policy, $\pi_T(\xt, \dt)$, as a composition of two modules $g_{\theta_{d}}$ and $\pi_{\theta_{b}}$, such that $\pi_T(\xt, \dt) = \pi_{\theta_{b}}(\xt, g_{\theta_{d}}(\dt))$. The first module $g_{\theta_{d}}$ is the encoder,
\begin{equation}
\textbf{z}_t = g_{\theta_{d}}(\dt)\,,
\end{equation}
which compresses $\dt$ into an intermediate latent vector $\zt$. The second module $\pi_{\theta_{b}}$ is the policy body,
\begin{equation}
\at = \pi_{\theta_{b}}(\xt, \zt)\,,
\end{equation}
which predicts an action from the latent $\textbf{z}_t$ and observation $\xt$. Each module is parameterized as a neural network with ELU activations and architecture described in Table \ref{tab:network_arch}. We optimize the teacher's parameters $\theta_{d}, \theta_{b}$ together using PPO \cite{schulman2017proximal} to maximize the future discounted reward,

\begin{equation}
\max_{\theta_{b}, \theta_{e}} \mathbb{E}_{\pi_{\theta_{b}, \theta_{e}}}\big[\sum\limits_{t=0}^\infty \gamma^t r_t\big]\,.
\end{equation}

\begin{table}[t!]
\centering\small
\begin{tabular}{lccc}
\toprule
           Module      &  Inputs &    Hidden Layers &    Outputs   \\
\midrule

Enc.   ($g_{\theta_{d}}$)      & $\textbf{d}_{t}$ ($12$) &  [256, 128]  & $\textbf{z}_t$ (8)   \\
Adapt.   ($h_{\theta_{a}}$)            & $\textbf{x}_{[t-h:t-1]}$ ($42\times15$) & [256, 32] & $\textbf{z}_t$ (8)  \\
Body ($\pi_{\theta_{b}}$)           & $\xt$ (42), $\textbf{z}_t$ (8)  & [512, 256, 128] & $\at$ (12)  \\
\bottomrule
\end{tabular}
\caption{Network architecture for encoder $g_{\theta_{d}}$, adaptation module $h_{\theta_{a}}$, and policy body $\pi_{\theta_{b}}$. The teacher policy is ${\pi_T(\xt, \dt) = \pi_{\theta_{b}}(\xt, g_{\theta_{d}}(\dt))}$, with parameters $\theta_{b}, \theta_{d}$ optimized using PPO. The student policy, ${\pi_S(\xt, \textbf{x}_{[t-h:t-1]}) = \pi_{\theta_{b}}(\xt, h_{\theta_{a}}( \textbf{x}_{[t-h:t-1]}))}$ reuses $\theta_{b}$ from the teacher and $\theta_{a}$ is optimized using supervised learning. }\label{tab:network_arch}
\end{table}

\subsubsection{Student Policy}
The student policy $\pi_S(\xt, \textbf{x}_{[t-h:t-1]}) = \pi_{\theta_{b}}(\xt, h_{\theta_{a}}( \textbf{x}_{[t-h:t-1]}))$ imitates the teacher's behavior during deployment without access to $\dt$.
The student policy is constructed by replacing encoder $g_{\theta_{d}}(\dt)$ with an online identification module~\cite{kumar2021rapid, lee2020learning},
\begin{equation}
\hat{\textbf{z}}_t = h_{\theta_{a}}(\textbf{x}_{[t-h:t-1]})\,,
\end{equation}
which estimates the latent $\zht$ from state history $\textbf{x}_{[t-h:t-1]}$. We train the identification module so that its predictions $\zht$ match the encoder's output $\zt = g_{\theta_{d}}(\dt)$ as closely as possible. To this end, we optimize parameters $\theta_{a}$ using supervised learning on on-policy data, using the loss function
\begin{equation}
\mathcal{L}_{\theta_{d}} = \Big(h_{\theta_{a}}(\textbf{x}_{[t-h:t-1]}) -  g_{\theta_{d}}\big(\dt)\Big)^2 = (\zht - \zt)^2\,.
\end{equation}

When this loss is low, the latent representation $\zt$ is shared between the teacher and the student, so the student can reuse the teacher's policy body module as $\at = \pi_{\theta_{b}}(\zht, \xt)$ to select actions without further training. 

The optimization procedure closely follows that of~\cite{kumar2021rapid, lee2020learning} with a few minor differences: (1) We use a shorter history of \(h=15\) observations, small enough for the adaptation module to run in real-time synchronously with the policy body; (2) We train the adaptation module simultaneously with the teacher using on-policy data. 
We found that the robot's ability to run at high speed was not sensitive to these design choices, which made training and deployment easier. Table \ref{tab:network_arch} gives the architecture of each component of the system. PPO hyperparameters are listed in Table \ref{tbl:hyperparams} (Appendix).

\subsection{Curriculum Strategy}
\label{sec:method_curriculum}
The agent learns a velocity-conditioned policy by attempting to track different velocity commands during training.
To this end, the longitudinal and yaw velocity commands $\textbf{v}^{\footnotesize{\textrm{cmd}}}_x$, $\boldsymbol{\omega}^{\footnotesize{\textrm{cmd}}}_z$ during episode $k$ are sampled from a probability distribution $p^k_{\textbf{v}_x, \boldsymbol{\omega}_z}(\cdot, \cdot)$. The lateral velocity command $\textbf{v}^{\footnotesize{\textrm{cmd}}}_y$ is sampled separately from a small uniform probability distribution because longitudinal and yaw speed are sufficient for omnidirectional locomotion. 
Without a curriculum, there is no change in the sampling procedure from episode to episode:

\begin{equation}
p^{k+1}_{\textbf{v}_x, \boldsymbol{\omega}_z}(\cdot, \cdot) \leftarrow p^{k}_{\textbf{v}_x, \boldsymbol{\omega}_z}(\cdot, \cdot)\,.
\end{equation}

When velocity commands are sampled uniformly from a small range ($\textbf{v}^{\footnotesize{\textrm{cmd}}}_x \in [-1.0, 1.0]$, $\boldsymbol{\omega}^{\footnotesize{\textrm{cmd}}}_z \in [-1.0, 1.0]$) at the start of training, the agent can learn to track them \cite{hwangbo2019learning,kumar2021rapid}. However, when commanded velocities are sampled uniformly from a large distribution ($\textbf{v}^{\footnotesize{\textrm{cmd}}}_x \in [-4.0, 4.0]$, $\boldsymbol{\omega}^{\footnotesize{\textrm{cmd}}}_z \in [-5.0, 5.0]$), we found that learning fails (Figure \ref{fig:curriculum}).

The reason for failure is that locomotion at high speeds is challenging, and if most of the commands are high-velocity, the agent fails to gather enough reward. This problem may be mitigated if we first expose the agent to low-velocity commands and gradually increase the desired speed 
via a curriculum
~\cite{bengio2009curriculum}). 
Some works use a curriculum
where the commands are updated on a fixed schedule, as a function of the timing variable $k$. This update rule $f$ takes the form: 

\begin{equation}
p^{k+1}_{\textbf{v}_x, \boldsymbol{\omega}_z}(\cdot, \cdot) \leftarrow f\big(p^{k}_{\textbf{v}_x, \boldsymbol{\omega}_z}(\cdot, \cdot), k\big)\,.
\end{equation}

A fixed schedule requires manual tuning. Moreover, if the system designer modifies the environment or learning algorithm, the agent's learning speed will be different, which would necessitate re-tuning the curriculum schedule. 
Rather than advancing the command curriculum on a fixed schedule, we automatically update the curriculum using a reward-based rule~\cite{akkaya2019solving, li2020towards, matiisen2019teacher, xie2020allsteps}.
One possibility is to maintain independent distributions over command dimensions  $p_{\textbf{v}_x}(\cdot), p_{\boldsymbol{\omega}_z}(\cdot)$ such that $p_{\textbf{v}_x, \boldsymbol{\omega}_z}(\cdot, \cdot) = p_{\textbf{v}_x}(\cdot) p_{\boldsymbol{\omega}_z}(\cdot)$, and to specify the update rules $f_v$, $f_\omega$ for each component separately:

\begin{figure*}[t!]
    \centering
    \begin{subfigure}[t]{0.45\textwidth}
        \centering
        \includegraphics[height=105pt]{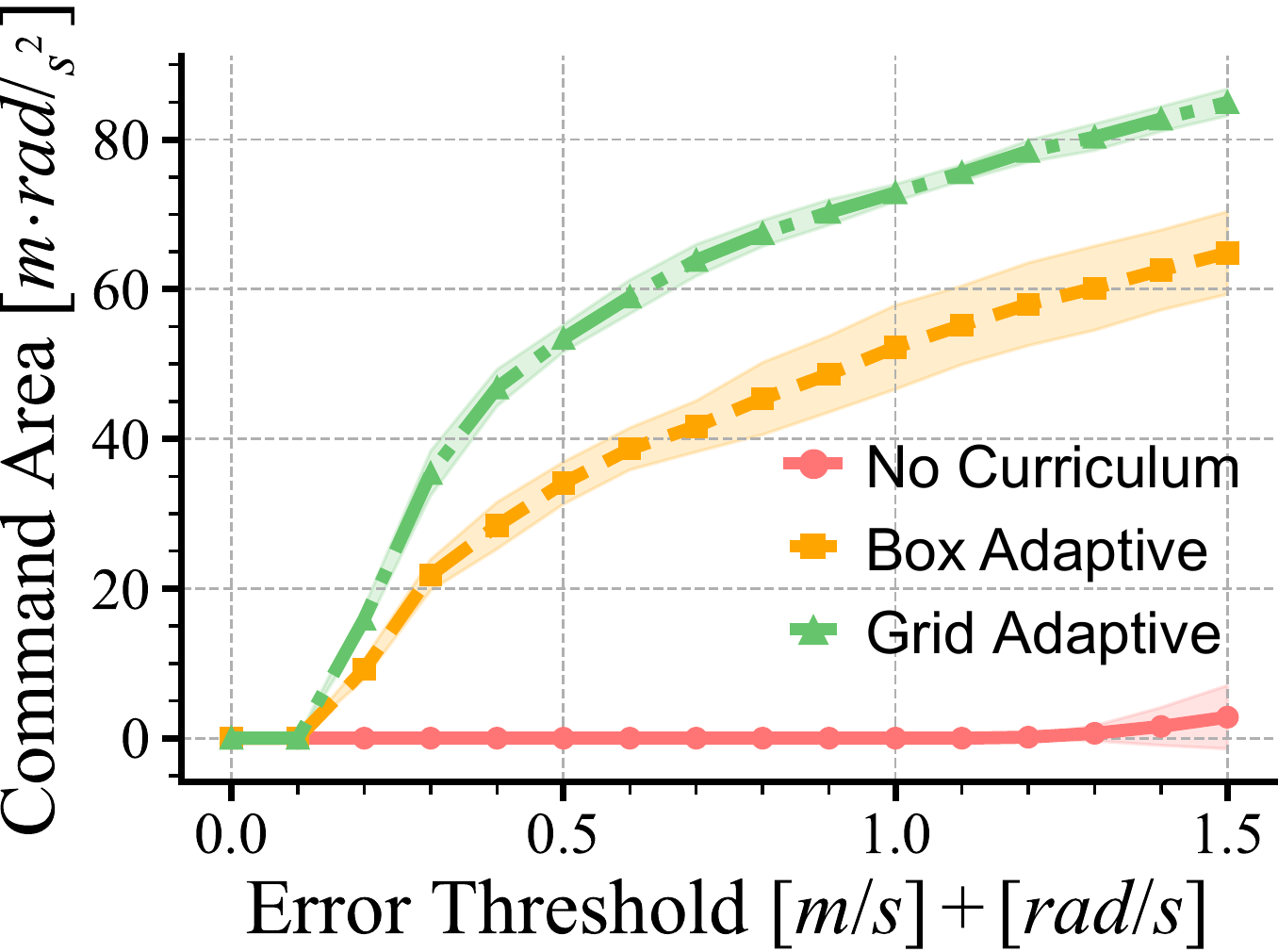}
        \caption{Command area vs. error threshold}\label{fig:curr_bar}
    \end{subfigure}
    \begin{subfigure}[t]{0.50\textwidth}
        \centering
        \includegraphics[height=105pt]{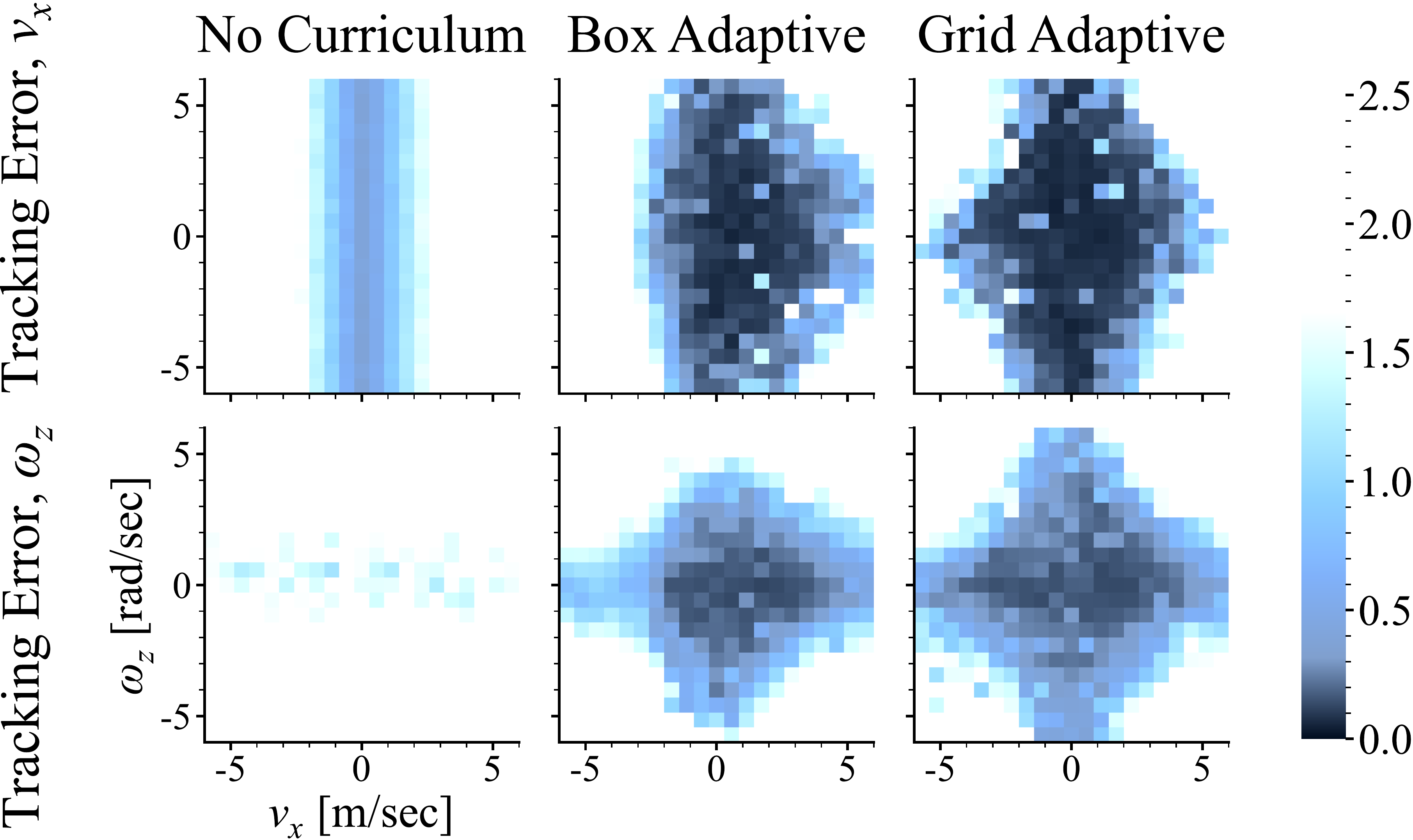}
        \caption{Heatmap of converged tracking error for curricular strategies.}\label{fig:curr_heatmap}
    \end{subfigure}\hfill
    \caption{
   (a) Forward and angular velocity tracking performance. The Grid Adaptive curriculum tracks a larger range of velocities than the Box Adaptive curriculum for all error thresholds. (b) Velocity tracking error in the forward axis (top) and yaw axis (bottom); darker is better. 
   In each heatmap, the x-axis varies the forward velocity command between \([-6, \SI{6}{\meter/\second}] \) and the y-axis varies the yaw rate between \([-6, \SI{6}{\radian/\second}]\).
   From left to right: \textit{\color{red} No Curriculum} fails to learn meaningful velocity control; its heatmaps correspond to a robot jittering in place, as its tracking error is equal to the command. \textit{\color{orange} Box Adaptive} curriculum learns to control the robot but excludes extremes of the command space. \textit{\color{black!60!green} Grid Adaptive} curriculum covers a larger command area by accounting for the combined impact of running and turning speed on task difficulty.
   }\label{fig:curriculum}
\end{figure*}

\begin{subequations}
\begin{equation}
p^{k+1}_{\textbf{v}_x}(\cdot) \leftarrow f_v\big(p^{k}_{\textbf{v}_x}(\cdot), r_{v_x}\big)\,,
\end{equation}
\begin{equation}
p^{k+1}_{\boldsymbol{\omega}_z}(\cdot) \leftarrow f_\omega\big(p^{k}_{\boldsymbol{\omega}_z}(\cdot), r_{\omega_z}\big)\,,
\end{equation}
\end{subequations}
where $r_{v_x}$ and $r_{\omega_z}$ are the velocity tracking rewards as detailed in Table \ref{tbl:rewards}. We refer to this approach as the Box Adaptive curriculum because the probability density function in the $\textbf{v}_x$-$\boldsymbol{\omega}_z$ plane is shaped like a box.

If $\textbf{v}_x$ and $\boldsymbol{\omega}_z$ are chosen independently, 
then commands with both high linear and angular velocity will be sampled equally as often as commands with just one large velocity component.
However, due to the effects of centrifugal force at high speeds, simultaneous running and turning are much more demanding than fast straight-line running or spinning in place.
A curriculum that independently increases the linear and angular velocities might fail to discover some behaviors because most high-speed commands are infeasible.
This motivates us to use a curriculum strategy that models the joint distribution over linear and angular velocity commands:

\begin{equation}
p^{k+1}_{\textbf{v}_x, \boldsymbol{\omega}_z}(\cdot, \cdot) \leftarrow f\big(p^{k}_{\textbf{v}_x, \boldsymbol{\omega}_z}(\cdot, \cdot), r_{v_x}, r_{\omega_z}\big)\,.
\end{equation}
We refer to this as the Grid Adaptive curriculum. 

Having described the form for the two curriculum strategies, we will now provide the detailed update rules. For both strategies we initialize $p^{k}_{\textbf{v}_x, \boldsymbol{\omega}_z}(\cdot, \cdot)$ as a uniform probability distribution over (${\vxcmd \in [-1.0, 1.0]}$, ${\wzcmd \in [-1.0, 1.0]}$). We represent distribution $p^{k}_{\textbf{v}_x, \boldsymbol{\omega}_z}(\cdot, \cdot)$ as a discrete grid with resolution $[\SI{0.5}{\meter/\second}, \SI{0.5}{\radian/\second}]$ centered at $[\SI{0}{\meter/\second}, \SI{0}{\radian/\second}]$. To control the growth of the sampling distribution, we define a success threshold, $\gamma$, with constant value between 0 and 1. 

\subsubsection{Box Adaptive Curriculum Update Rule}
At episode $k$, the linear and angular velocity commands for the agent are sampled independently: $ \vxcmd \sim p_{\textbf{v}}^{k}(\cdot)$, $\wzcmd \sim p_{\boldsymbol{\omega}_z}^{k}(\cdot)$. If the agent succeeds in this region of command space, we would like to add neighboring regions to the sampling distribution, potentially increasing the difficulty of future commands.
Suppose the agent receives rewards ${\rvx, \rwz}$ in its attempt to follow $\vxcmd$, $\wzcmd$. Then we apply the update rule

\begin{subequations}
\begin{equation}
p_{\textbf{v}_x}^{k+1}(\textbf{v}^{\textrm{n}}_x) \leftarrow  
\begin{cases} 
      p_{\textbf{v}_x}^{k}(\textbf{v}^{\textrm{n}}_x) & r_{v^{\footnotesize{\textrm{cmd}}}_x} < \gamma\,, \\
      1 & \text{otherwise.} 
   \end{cases}
\end{equation}
\begin{equation}
p_{\boldsymbol{\omega}_z}^{k+1}(\boldsymbol{\omega}^{\textrm{n}}_z) \leftarrow  
\begin{cases} 
      p_{\boldsymbol{\omega}_z}^{k}(\boldsymbol{\omega}^{\textrm{n}}_z) & r_{\omega^{\footnotesize{\textrm{cmd}}}_z} < \gamma\,, \\
      1 & \text{otherwise.} 
   \end{cases}
\end{equation}
\end{subequations}
which increases the probability density on neighbors $\textbf{v}^{ \textrm{n}}_x$ of $\textbf{v}^{\footnotesize{\textrm{cmd}}}_x$ and $\boldsymbol{\omega}^{\textrm{n}}_z$ of $\boldsymbol{\omega}^{\footnotesize{\textrm{cmd}}}_z$.
Here, neighboring commands are defined as the adjacent elements in the (discretized) domain of each marginal distribution: $\textbf{v}^\textrm{n}_x \in \{\vxcmd-0.5, \vxcmd+0.5\}$ and $\boldsymbol{\omega}^{\textrm{n}}_z \in \{\wzcmd-0.5, \wzcmd+0.5\}$.
Suppose $\textbf{v}^{\footnotesize{\textrm{cmd}}}_x$ or $\boldsymbol{\omega}^{\footnotesize{\textrm{cmd}}}_z$ is among the most challenging commands in one of the distributions, and the reward threshold is met. In that case, this update will result in that distribution expanding.

\subsubsection{Grid Adaptive Curriculum Update Rule}
At episode $k$, the linear and angular velocity commands for the agent are sampled from the joint distribution: $ \textbf{v}^{\footnotesize{\textrm{cmd}}}_x , \boldsymbol{\omega}^{\footnotesize{\textrm{cmd}}}_z \sim p^{k}_{\textbf{v}_x, \boldsymbol{\omega}_z}(\cdot, \cdot)$.
As before, if the agent succeeds in this region of command space, we would like to increase the difficulty by adding neighboring regions to the sampling distribution. However, the distributions of $\vxcmd$ and $\vycmd$ are no longer constrained to be independent. This enables us to revise our update with a new definition of the neighboring commands.
Upon termination of an episode with command ${[\textbf{v}^{\footnotesize{\textrm{cmd}}}_x, \boldsymbol{\omega}^{\footnotesize{\textrm{cmd}}}_z]}$ where the agent received rewards ${r_{v^{\footnotesize{\textrm{cmd}}}_x}, r_{\omega^{\footnotesize{\textrm{cmd}}}_z}}$, we use the following update:

\begin{equation}
p^{k+1}_{\textbf{v}_x, \boldsymbol{\omega}_z}(\textbf{v}^{\textrm{n}}_x, \boldsymbol{\omega}^{\textrm{n}}_z) 
\leftarrow   \begin{cases} 
      p^{k}_{\textbf{v}_x, \boldsymbol{\omega}_z}(\textbf{v}^{\textrm{n}}_x, \boldsymbol{\omega}^{\textrm{n}}_z) & r_{v^{\footnotesize{\textrm{cmd}}}_x} < \gamma \textrm{ or } r_{\omega^{\footnotesize{\textrm{cmd}}}_z} < \gamma, \\
      1 & \text{otherwise.}
   \end{cases}
\end{equation}

This update adds probability density to the neighboring velocity commands $[\textbf{v}^\textrm{n}_x, \boldsymbol{\omega}^\textrm{n}_z]$ of $[\textbf{v}^{\footnotesize{\textrm{cmd}}}_x, \boldsymbol{\omega}^{\footnotesize{\textrm{cmd}}}_z]$, if those commands have not already been added. Here, neighboring commands are defined as neighbors in the 4-connected grid domain of $p^{k}_{\textbf{v}_x, \boldsymbol{\omega}_z}(\cdot, \cdot)$, which is a discrete grid with resolution $[\SI{0.5}{\meter/\second}, \SI{0.5}{\radian/\second}]$. If $[\textbf{v}^{\footnotesize{\textrm{cmd}}}_x, \boldsymbol{\omega}^{\footnotesize{\textrm{cmd}}}_z]$ is among the most challenging commands in the joint distribution, and the reward threshold is met, this update will result in the distribution expanding locally. 

\subsection{Evaluation Metrics}
\label{sec:metrics}

The controller is tasked to track body velocity commands. Consider a command:  (\(\vxcmd, \wzcmd\)) corresponding to a point in the \(\vxcmd\)-\(\wzcmd\) plane. We discretize this plane into a grid with resolution [\SI{0.5}{\meter/\second}, \SI{0.5}{\radian/\second}] with grid cell indices denoted as $i, j$. Then, for each grid cell, we define the tracking error \(\epsilon_{ij}\) as the root mean square deviation, averaged over trials in that grid cell:

\begin{subequations}
\begin{equation}
\epsilon_{ij}[\textbf{v}_x^{\textrm{\footnotesize cmd}}] = \mathbb{E}_{\textbf{v}_x^{\textrm{\tiny cmd}} \sim [i-1, i],\boldsymbol{\omega}_z^{\textrm{\tiny cmd}} \sim [j-1, j]} \sqrt{\mathbb{E}_{t} (\textbf{v}_x^{\textrm{\tiny cmd}} - \textbf{v}_x^t)^2}\,,
\end{equation}
\begin{equation}
\epsilon_{ij}[\boldsymbol{\omega}_z^{\textrm{\footnotesize cmd}}] = \mathbb{E}_{\textbf{v}_x^{\textrm{\tiny cmd}} \sim [i-1, i], \boldsymbol{\omega}_z^{\textrm{\tiny cmd}} \sim [j-1, j]} \sqrt{\mathbb{E}_{t} (\boldsymbol{\omega}_z^{\textrm{\tiny cmd}} - \boldsymbol{\omega}_z^t)^2}\,,
\end{equation}
\end{subequations}
where \(\textbf{v}_x^t, \boldsymbol{\omega}_z^t\) are the forward and yaw velocity of the robot measured at time $t$. In our experiments, we compute tracking error from 5 trials per grid cell.

Measuring either the longitudinal or yaw velocity in isolation does not provide a complete picture of controller performance. Instead, we want a metric that captures the combinations of longitudinal and yaw velocity that the robot is able to track. To this end, we constructed an aggregate metric that captures the diversity of commands the controller can actuate given some maximum error tolerance. For a certain error threshold \(\epsilon_0\), we define the \textit{command area} as the area of the region in the \(\textbf{v}_x^{\textrm{\footnotesize cmd}}\)-\(\boldsymbol{\omega}_z^{\textrm{\footnotesize cmd}}\) plane for which the tracking errors satisfy 
\begin{equation}
\epsilon_{ij}[\textbf{v}_x^{\textrm{\footnotesize cmd}}] + \epsilon_{ij}[\boldsymbol{\omega}_z^{\textrm{\footnotesize cmd}}] < \epsilon_0.
\end{equation}

The dimension of the command area is $\SI{}{\meter / \second} \cdot \SI{}{\radian/\second}$. 
Intuitively, if one controller has a larger command area than another, the former can achieve a greater range of speeds while remaining below the same error threshold \(\epsilon_0\). When we report the command area, we evaluate policies trained with five random seeds and indicate their standard deviation using an error bar.

\section{Results}
\label{sec:results}

\subsection{Curriculum Learning Enables High-Speed Locomotion}
\label{sec:results_curr}

Figure \ref{fig:curriculum}(a) visualizes the tracking error (see Section~\ref{sec:metrics}) of the policies learned from the three command sampling strategies as heatmaps in the $\vxcmd$-$\wzcmd$ plane. The shading on each heatmap corresponds to tracking error, with darker shades indicating lower error. We observe that the policy trained without any curriculum fails to learn. This is because the robot's random exploration at the start of training rarely results in fast body motion. Hence, the reward almost always remains small, providing minimal learning signal. 

The performance of the system is improved substantially by implementing the Box Curriculum. The agent first learns to track well in the small initial command distribution, then gradually increases its capability as the commands become larger. 

Using the Grid Curriculum, the performance of the policy further improves, as evidenced by the larger command area. It achieves this by maintaining a full joint distribution over linear and angular velocity, thereby modeling their interaction. 
When high linear and angular velocities are combined, a body experiences a centrifugal force which must be countered by frictional force to remain on the desired path. This force balance induces a constraint on maximum combinations of linear and angular velocity such that the two vary inversely $[\boldsymbol\omega_z \sim 1/\textbf{v}_x]$ when the constraint is active. 
This phenomenon is in agreement with the apparent inverse shape of the command area boundary shown in Figure \ref{fig:curr_heatmap}, which suggests that the robot has reached a physical limit on its ability to turn at high speed.
The Grid Curriculum can limit itself to sampling combinations of linear and angular speed that are jointly feasible.
In contrast, because the Box Curriculum samples linear and angular velocity independently, it will frequently 
generate infeasible high-speed tasks that hinder learning.

\subsection{Real-world Testing}

\begin{table}[t!]
\centering\small
\begin{tabular}{llcccc}
\toprule
                                 & Robot        & RL?      & Froude  & Speed   & Leg $L$ \\
                                 &              &            & (-)  & \SI{}{\meter/\second} & \SI{}{\cm} \\
\midrule
\citeauthor{park2017high}        & C2        &           & 7.1 & 6.4 & 59 \\
\textbf{Ours}, \citeauthor{ji2022concurrent}                 & MC     & \textbf{Y}    & \textbf{5.1} & \textbf{3.9} & 30 \\
\citeauthor{kim2019highly}     & MC     &           & 4.6 & 3.7 & 30 \\

Unitree                     & A1           &           & 2.8  & 3.3 & 40 \\
\citeauthor{kumar2021rapid}        & A1           & Y    & 0.8 & 1.8  & 40 \\
\footnotesize{\citeauthor{hwangbo2019learning}} & ANYmal       & Y    & 0.5 & 1.5  & 50 \\
\bottomrule
\end{tabular}
\caption{Measure of Agility: Comparison between the Froude numbers of various prior works. MC = Mini Cheetah; C2 = Cheetah 2. We are the first to demonstrate that reinforcement learning (RL) achieves agile locomotion with Froude number $\geq 1$ (along with concurrent work \citeauthor{ji2022concurrent}). } 
\label{tab:froude}
\end{table}

Video of all experiments described in this section is viewable on the project website: \url{https://agility.csail.mit.edu/}.

\textbf{Indoor Running} To evaluate how fast our robot can run in the real world, we ramped the velocity command to \SI{6.0}{\meter/\second}. We conducted this experiment in a motion capture arena to accurately estimate the robot's running speed (Figure \ref{fig:teaser}, top). We found that policies trained with a system identification module and grid curriculum sustained an average speed of \SI{3.8}{\meter/\second} across multiple seeds (Table \ref{tab:sim_to_real}), with the highest sustained speed of \SI{3.9}{\meter/\second} among the three seeds. This is higher than the previous record of \SI{3.7}{\meter/\second} reported for a model-predictive control algorithm on the same robot~\cite{kim2019highly}. Together with concurrent work \cite{ji2022concurrent}, this is substantially faster than previous applications of RL to legged locomotion.

The maximum attainable speed is intimately tied to the robot's hardware properties, such as its weight, motor strength, and leg length. Although there is no perfect way to compare agility across different robot designs, \textit{Froude number}~\cite{alexander1984gaits} normalizes a robot's speed by its leg length and has been used to measure agility across robot platforms in the past~\cite{park2017high}. 
Table~\ref{tab:froude} compares the Froude numbers across different quadrupeds and controllers. Along with the concurrent work of~\citet{ji2022concurrent}, our method is substantially more agile than previous applications of reinforcement learning. 

While our robot successfully performs rapid locomotion in the real world, there exists a sim-to-real gap as reported in Table~\ref{tab:sim_to_real}. The results reveal that online system identification leads to better tracking of the velocity command of \SI{6.0}{\meter/\second} in simulation  (speed of \SI{5.46}{\meter/\second} with and \SI{5.07} {\meter/\second} without system identification), and also reduces the sim-to-real gap (average speeds of \SI{3.81} {\meter/\second} with and \SI{2.49} {\meter/\second} without system identification). While prior work demonstrated that sim-to-real gap can be mitigated at low velocities~\cite{kumar2021rapid,lee2020learning}, our results show that these findings also hold true at high speeds. 

Some of the remaining sim-to-real performance gap may result from an inaccurate selection of simulated terrain and robot parameters for evaluation. For example, the configuration used for simulated evaluation might overestimate the robot's effective motor strength or the ground friction, resulting in a different maximum speed. On the other hand, some aspects of the real-world dynamics are probably not captured under any configuration of the simulator. This type of sim-to-real gap could result in suboptimal real-world top speed. The relative contributions of these factors to the sim-to-real performance gap remains uncertain.

\label{sec:real-world-performance}

\begin{table}[t!]
\centering\small
\begin{tabular}{llll}
\toprule
                                        & $\textbf{v}^{\textrm{cmd}}_x$ & $\textbf{v}_x$ (Sim) & $\textbf{v}_x$ (Real)  \\
\midrule
With System ID ($\pi_{\theta_{ST}}$) & \SI{6.0 }{\meter/\second} &  5.46  &  $3.81 \pm 0.09$  (3) \\
W/o System ID ($\pi_{\theta_{DR}}$) &  \SI{6.0 }{\meter/\second} &  5.07 &  $ 2.49 \pm 0.07$ (2) \\
\bottomrule
\end{tabular}
\caption{Quantifying the sim-to-real gap in the maximum-velocity regime. Velocity on the real robot is measured using a precise motion capture setup.}
\label{tab:sim_to_real}
\end{table}

\textbf{Outdoor Running} Outdoor terrain presents multiple challenges not present in indoor running, among which are changes in ground height, friction, and terrain deformation. Under these variations, the robot must actuate its joints differently to reach high speed than it would on flat, rigid terrain with high friction, such as a treadmill or paved road. 
To test if our system can run on outdoor terrains, we conducted an outdoor dash across an uneven grassy patch as shown in Figure \ref{fig:teaser} (second row). We record an outdoor \(10\)-meter dash time of \(2.94\) seconds, corresponding to an average speed of \SI{3.4}{\meter/\second}. 

\textbf{Yaw Control}
We evaluate our controller's yaw velocity control in the lab setting as shown in Figure \ref{fig:teaser} (third row). The robot accelerates to a maximum yaw rate of \SI{5.7}{\radian/\second}, then stops safely. This is \(90\%\) of the fastest yaw rate recorded on the Mini Cheetah using a model-based controller, at \SI{6.28}{\radian/\second} \cite{bledt2018cheetah}. However, the model-based records were achieved using two different controllers for linear~\cite{kim2019highly} and angular~\cite{bledt2018cheetah} velocity.  In contrast, a single policy achieved all indoor and outdoor running and spinning results in our work. To challenge the controller's spinning skills, we brought the robot outside after a snowstorm and piloted it onto an icy patch, illustrated in Figure \ref{fig:teaser} (bottom). The robot maintained stability while spinning as its feet frequently slipped on ice.

\textbf{Response to Terrain Changes and Hardware Failures}
We tested our system in a diverse set of challenging real-world scenarios: (1) ascending a steep incline made of small pebbles. (2) maintaining balance despite a mechanical blockage to one motor. (3) tripping at high speed, flying upside down, and landing on its feet. (4) recovering via a change in gait after tripping over a small barrier. We present these qualitative results in the accompanying video.

We also deployed the model-predictive controller from \cite{kim2019highly} in scenarios (1) and (4), which were the most convenient to replicate. Unlike our learned controller, the baseline did not recover from (1) slipping down the gravelly incline and (4) tripping over the barrier. While these results highlight the robustness of policies, we want to emphasize that we are not claiming that such (or even more) robustness cannot be achieved with model-predictive control. 
Our claim is simply that by freeing the human from the tedious task of refining the robot's model or behavior, the RL paradigm offers a scalable alternative to obtain robust behavior in diverse conditions.

\subsection{Ablation Studies}
\label{sec:ablation}

\subsubsection{Impact of Online System Identification}

\begin{figure}[t]
    \centering
        \includegraphics[width=0.6\linewidth]{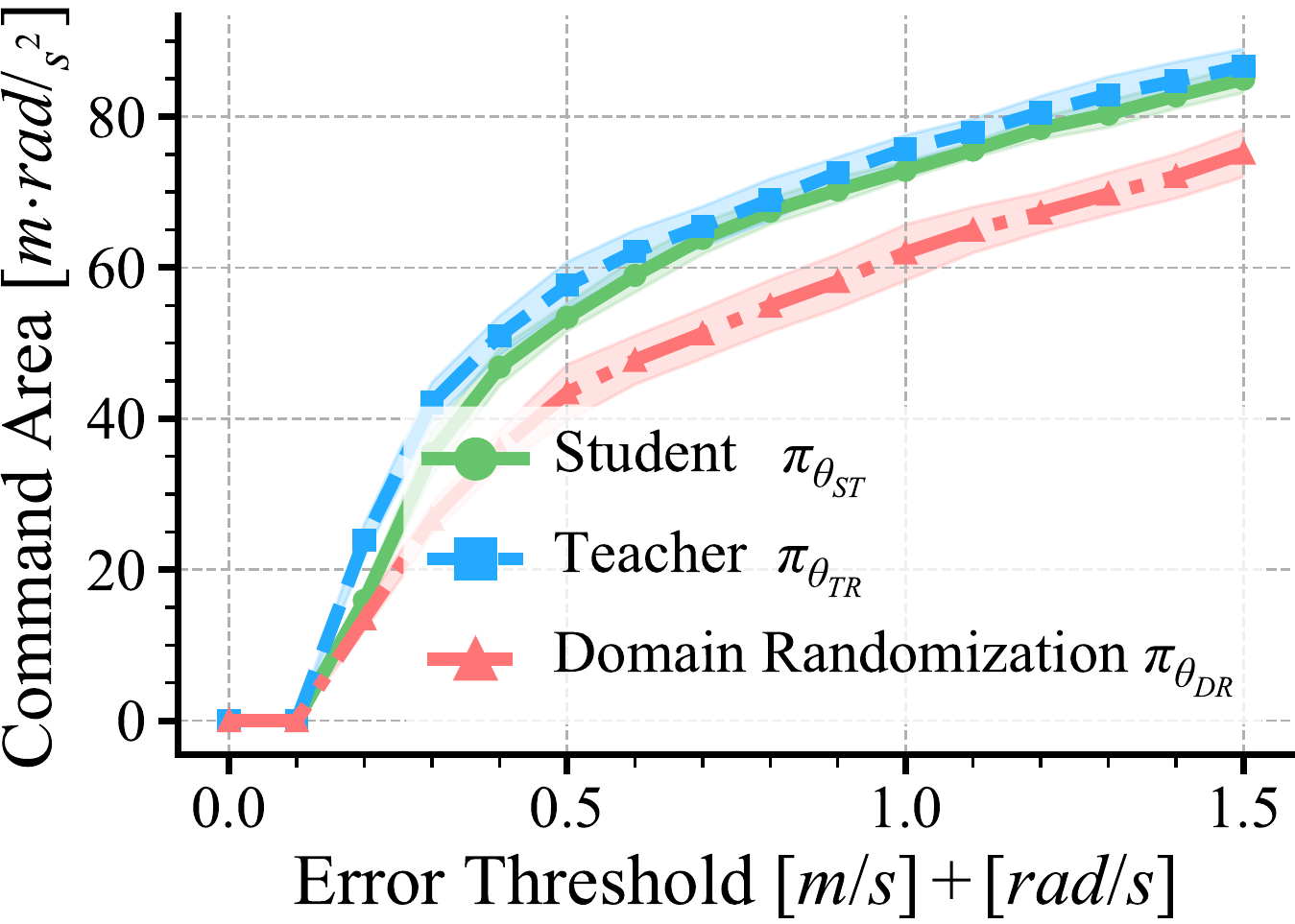}
    \caption{Online system identification reduces tracking error, particularly at high speeds. The command area increases as the error threshold is relaxed for teacher, student, and domain-randomized policies. 
    }\label{fig:privileged_info}
\end{figure}

System identification can become both more critical and more challenging as locomotion speed increases; this has been previously suggested by studies of model-based control systems~\cite{bosworth2016robot, fahmi2020stance}, but has not been explored in the context of reinforcement learning. We evaluate this hypothesis in the teacher-student setting by quantifying 
(1) the benefit of access to privileged information when learning to run at high speeds, and (2) the ability of the student policy to retain the performance gains using only available sensor data. We compare teacher, student, and domain randomized policies
as described in Section \ref{sec:method_opt} in the high-speed regime. All policies are trained under the same randomization of the privileged state (Table \ref{tab:domain_rand}).

We find that access to privileged information yields increased performance across all speeds, with the greatest benefit at high speeds. Figure \ref{fig:privileged_info} plots the command area (Section \ref{sec:metrics}) for the three policies as the threshold for error increases. The privileged teacher $\pi_T$ trained with access to environment parameters attains a strictly larger command area than the policy $\pi_{DR}$ trained with only the robot state. Using the online system identification module, we show that the student policy $\pi_S$ can nearly match the teacher's performance. The student's ability to imitate the teacher is consistent across all speeds.

\subsubsection{Impact of Rough-Terrain Training}
\label{sec:rough-terrain-impact}

\begin{figure}
    \centering
    \captionsetup{position=top}
    \begin{subfigure}[b]{0.29\textwidth}
        \includegraphics[height=100pt]{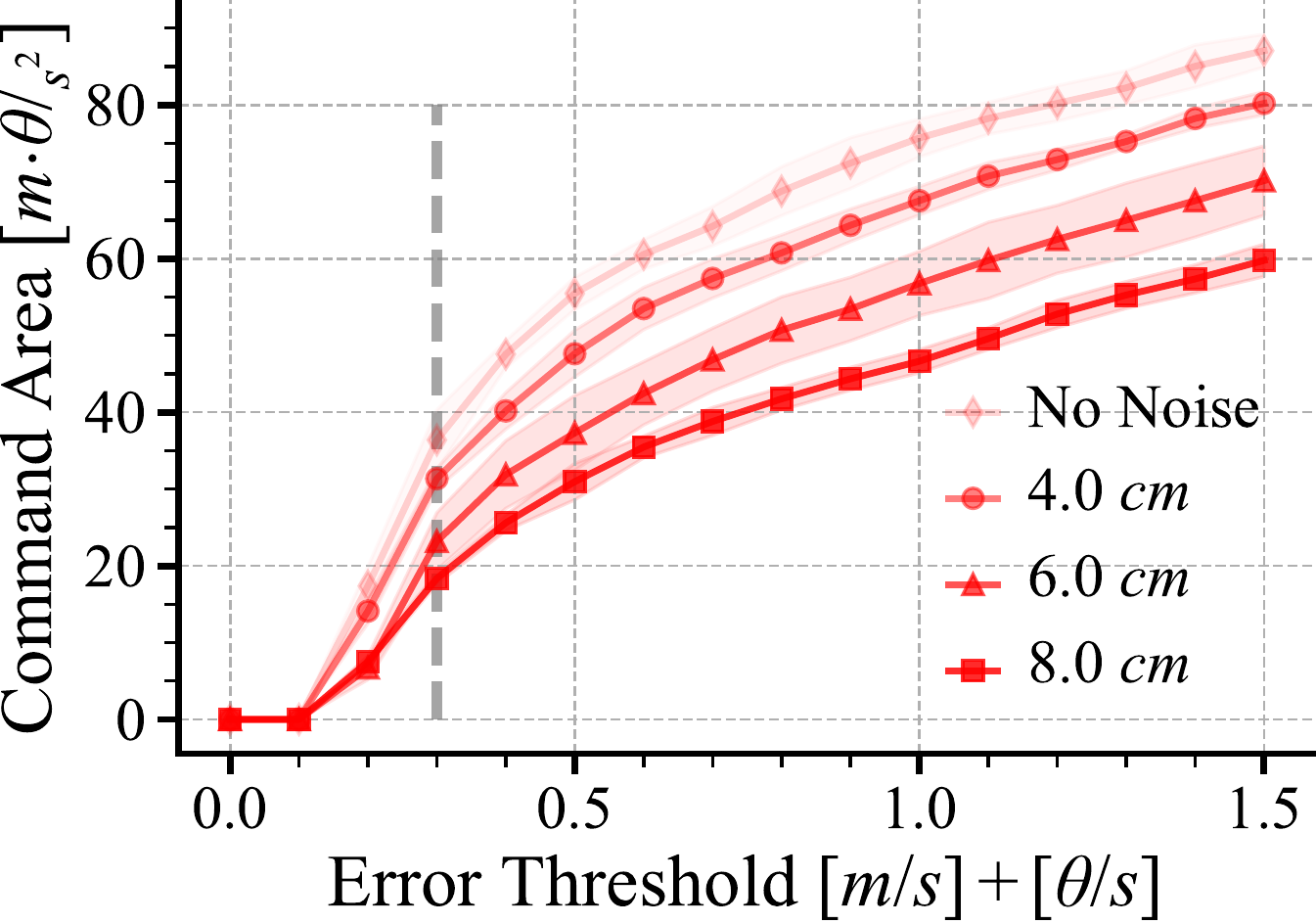}
        \caption{Command area vs. error threshold}\label{fig:terrain_noise_trend}
    \end{subfigure}
    \hfill
    \begin{subfigure}[b]{0.19\textwidth}
        \includegraphics[height=100pt]{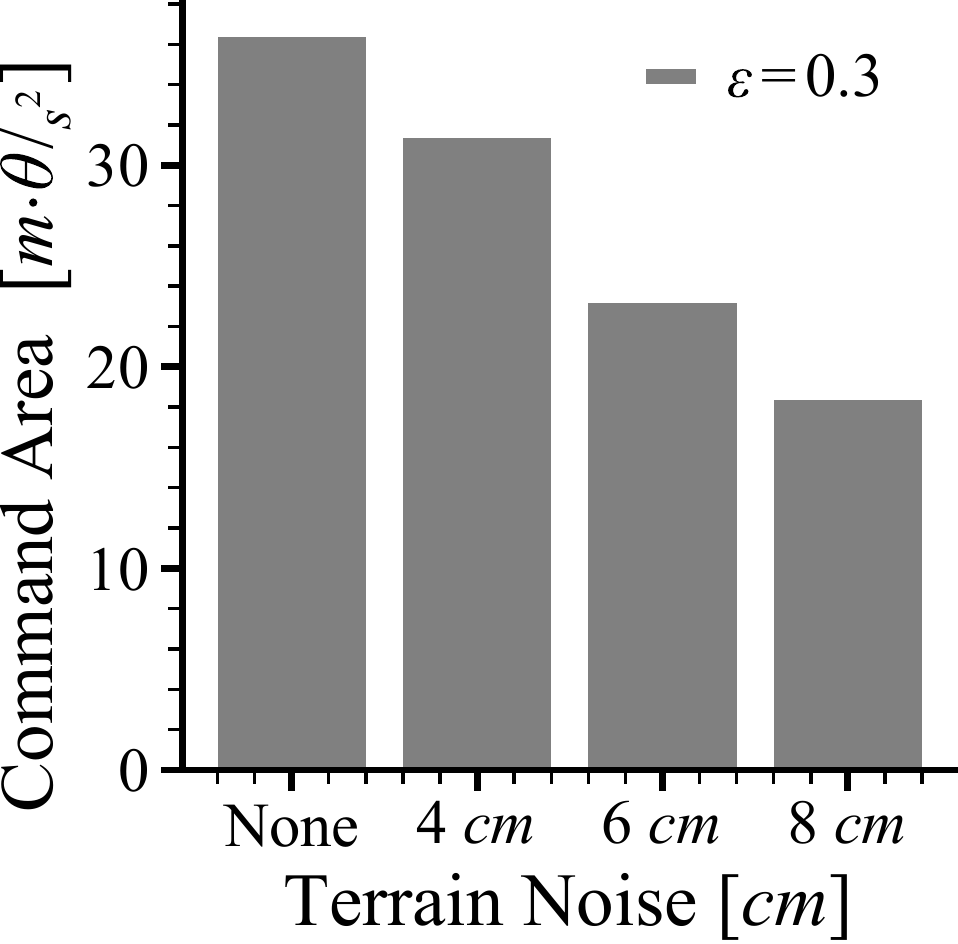}
        \caption{Slice along \(\epsilon=0.3\)}\label{fig:terrain_noise_bar}
    \end{subfigure}
    \caption{
    (a) Increasing the magnitude of terrain roughness during training shrinks the range of commands the robot can successfully track -- the command area -- on flat ground. This reflects a trade-off between robustness to rough terrains and top speed on flat ground. 
    (b) Histogram comparing the command area at error threshold $\epsilon = 0.3$, which corresponds to the {\color{gray}\textbf{gray}} vertical line on the left.} \label{fig:terrain_noising}
\end{figure}

One might hypothesize that for a system to operate on rough terrains, it must also be trained on rough terrains. The strategy of training on rough terrains has been applied successfully in prior works \cite{kumar2021rapid, lee2020learning, siekmann2021blind} to enable robust locomotion on diverse terrains. We find that despite training only on flat ground, our policy is sufficiently robust to deploy on various outdoor terrains. 
Moreover, for a blind policy, there is a trade-off between speed on flat ground and robustness on uneven terrain.
Figure \ref{fig:terrain_noising} reports the decrease in performance brought on by introducing terrain roughness when training a high-velocity locomotion policy.

\section{Related Work}

\textbf{Model-based Control for Locomotion} Seminal work in the field used simplified models and hand-specified gaits to make legged robots balance and move dynamically~\cite{herdt2010walking, kajita2003biped, raibert1986legged}. Subsequent works introduced expanded models with layered control architectures capable of operation on subsets of rough, soft, and slippery terrains~\cite{bosworth2016robot, fahmi2020stance, kuindersma2016optimization,park2017high, raibert2008bigdog, righetti2012quadratic}.

Recent innovations have addressed specific limitations of simple models with respect to high-speed running. Whole-body control~\cite{dai2014whole} enables simultaneous modeling of robot dynamics and kinematic constraints in real-time. By framing the whole-body control task as one of regulating ground reaction forces, \cite{kim2019highly} formulated a controller capable of running at speeds up to 3.7m/s on the Mini Cheetah. Regularized Predictive Control \cite{bledt2018cheetah} additionally applied learned heuristics within a model-based framework, which expanded the robot's ability to spin at high speeds and make tight cornering turns.

\textbf{Reinforcement Learning for Locomotion}
\cite{tan2018simtoreal} combined model-free reinforcement learning with dynamics randomization to learn fast trotting and bounding controllers for the Minitaur robot to move at a fixed speed and direction on flat ground. Extending this approach, \cite{hwangbo2019learning} trained a velocity-tracking controller for the ANYmal robot for speeds up to \SI{1.5}{\meter/\second}, and \cite{xie2020learning} applied sim-to-real reinforcement learning for agile locomotion on the Cassie biped. Followup works expanded ANYmal's robustness by training on diverse terrains using the teacher-student learning paradigm \cite{lee2020learning, miki2022learning, rudin2021learning}. The mechanical design of the ANYmal robot is thought to limit it from running at higher speeds. \cite{fu2021minimizing, kumar2021rapid} investigated the capability of model-free controllers to efficiently traverse diverse terrains on the Unitree A1, a small robot with similar size, actuation, and cost to the Mini Cheetah. Although the A1's built-in MPC controller has a maximum running speed of \SI{3.3}{\meter/\second}, these learning-based works only demonstrated the robot running up to maximum speed \SI{1.8}{\meter/\second}.  

Concurrently with our work, Ji et al. \cite{ji2022concurrent} also trained agile running policies for the Mini Cheetah robot using reinforcement learning. Unlike our work, \cite{ji2022concurrent} used a fixed-schedule curriculum on forward linear velocity only. In agreement with our results, \cite{ji2022concurrent} concluded that online system identification improves sim-to-real transfer in the high-speed locomotion setting.   Instead of learning as we did to implicitly adapt to different environments, \cite{ji2022concurrent} learned to explicitly estimate robot state components such as body velocity and contact probability.

\textbf{Curricula for On-Policy Reinforcement Learning} Prior works have shown that a curriculum on environments can enable the discovery of behaviors that are challenging to learn directly using reinforcement learning~\cite{bengio2009curriculum}. \cite{akkaya2019solving} demonstrated an Automatic Domain Randomization strategy in which domain randomization scales are increased based on agent performance. Curricula on environments have also been demonstrated in locomotion context; \cite{lee2020learning, miki2022learning, rudin2021learning} applied a curriculum on terrains to learn highly robust walking controllers on non-flat ground. \cite{xie2020allsteps} notably evaluated terrain curriculum strategies, including adaptive curricula, in the setting of stepping stone traversal with a physically simulated biped. 

\textbf{Student-Teacher Training} Learning with a privileged teacher has been leveraged for robotics in a number of previous works.  \cite{kumar2021rapid, lee2020learning} applied this approach to the task of blind walking. The teacher policy observed $\dt$, the dynamic properties of the robot and terrain, and the student learned to infer them from \(\textbf{x}_{[t-h:t-1]}\), the history of joint angles and IMU readings. \cite{margolis2021learning, miki2022learning} used the same approach to incorporate terrain geometric information into a locomotion policy. In these works, $\dt$ was a ground-truth geometric heightmap of the terrain. In \cite{miki2022learning}, the student policy observed a noisy heightmap. In \cite{margolis2021learning}, the student policy observed a forward-facing depth image. \cite{chen2021system} applied the teacher-student training approach to the task of object reorientation using a dexterous five-fingered hand. In this work, $\dt$ included the true position of the object as well as the ground-truth state of the hand's fingers. The student policy learned to imitate the teacher using point cloud observations and noisy joint angle readings that could be obtained in the real world.

\section{Discussion}
\label{sec:conclusion}

This work has shown that a neural network controller trained fully end-to-end in simulation can push a small quadruped to the limits of its agility, achieving omnidirectional mobility competitive with well-engineered model-predictive controllers in the regime of high speed. 
Because our controller uses minimal sensing, we can implement it on a low-cost robot~\cite{katz2019platform} with commercially available analogues~\cite{unitree2022website}. Therefore, our method can be readily tested and built upon by others using relatively accessible materials.

Our controller achieves a Froude number of \(5.1\), which is the highest reported on the Mini Cheetah but lower than the fastest published system, the Cheetah 2 ~\cite{park2017high}. The Froude number does not capture the impact of any mechanical differences between robots except leg length. We can only speculate whether the agility of alternative platforms comes from improved analytical controllers or hardware differences such as motor strength and weight distribution.

Instrumentation and repeatability limited our ability to characterize the robot's outdoor performance fully. We cannot use motion capture to record the robot's state outdoors as we do in the lab. Also, it is unsafe and impractical to record a large number of high-speed trips or flips on a real robot. This constrained our analysis of the robot's outdoor behavior to be more qualitative while we performed our quantitative analysis in the laboratory setting. 

The behaviors we demonstrate in this work are diverse but still limited relative to the full space of possible locomotion tasks. The system we demonstrate has only been trained to control the robot's body velocity in the ground plane. Other categories of behavior such as jumping, crouching, choreographed dance, and loco-manipulation were outside the scope of this work and would potentially require a very different task specification. Our system also does not use vision, so in general, it cannot perform tasks that require planning ahead of time, like efficiently ascending stairs or avoiding pitfalls. 

Finally, we emphasize that while our system demonstrates high speed, its distinctive locomotion gait should not be interpreted as generally ``better" than the many possible alternatives. On the contrary, many users of legged robots wish to optimize for objectives beyond speed, such as energy efficiency or minimization of wear on the robot. Body speed alone is an underspecified objective, meaning that there may be many equally preferable motions that attain the same speed.
Combining learned agile locomotion with additional specifications such as auxiliary objectives or human preferences remains a promising direction for future work.

\section*{Acknowledgment}
The authors thank
the members of the Improbable AI Lab and the Biomimetic Robotics Laboratory for providing valuable feedback on the project direction and the manuscript. We are grateful to MIT Supercloud and the Lincoln Laboratory Supercomputing Center for providing HPC resources. The Mini Cheetah robot used in this work was donated by the MIT Biomimetic Robotics Laboratory and NAVER. The Biomimetic Robotics Laboratory also provided hardware support for the robot. This research was supported by the DARPA Machine Common Sense Program, the MIT-IBM Watson AI Lab, and the National Science Foundation under Cooperative Agreement PHY-2019786 (The NSF AI Institute for Artificial Intelligence and Fundamental Interactions, \hyperlink{http://iaifi.org/}{http://iaifi.org/}). This research was also sponsored by the United States Air Force Research Laboratory and the United States Air Force Artificial Intelligence Accelerator and was accomplished under Cooperative Agreement Number FA8750-19-2-1000. The views and conclusions contained in this document are those of the authors and should not be interpreted as representing the official policies, either expressed or implied, of the United States Air Force or the U.S. Government. The U.S. Government is authorized to reproduce and distribute reprints for Government purposes, notwithstanding any copyright notation herein.

\bibliographystyle{plainnat}
\bibliography{references}

\newpage
\clearpage
\newpage

\begin{appendix}
\subsection{Training Parameters}

The PPO training parameters used for all experiments are provided in Table \ref{tbl:hyperparams}.

\begin{table}[t!]
    \centering
    \small
    \caption{Training Parameters}
    \begin{tabular}{cc}
    \toprule
    Hyperparameter & Value \\ [0.5ex]
    \midrule
     discount factor & 0.99 \\ [0.5ex]
     GAE parameter & 0.95 \\ [0.5ex]
      \# timesteps per rollout & 21 \\ [0.5ex]
      \# epochs per rollout & 5 \\ [0.5ex]
      \# minibatches per epoch & 4 \\ [0.5ex]
      entropy bonus ($\alpha_2$) & 0.01 \\ [0.5ex]
      value loss coefficient ($\alpha_1$) & 1.0 \\ [0.5ex]
      clip range & 0.2 \\ [0.5ex]
      reward normalization & yes \\ [0.5ex]
      learning rate & 1e-3 \\ [0.5ex]
      \# workers & 1 \\ [0.5ex]
      \# environments per worker & 4096 \\ [0.5ex]
      \# total timesteps & 400M \\ [0.5ex]
      optimizer & Adam \\ [0.5ex]
     \bottomrule
    \end{tabular}
    \label{tbl:hyperparams}
\end{table}
\subsection{Reward Function} The reward terms are provided in \ref{tbl:rewards}.

\subsection{Measures of Agility}\label{sec:measure}
Benchmarking the agility of legged robots cannot be accomplished by comparing speed alone due to differences in hardware. \cite{alexander1984gaits} proposed to characterize legged agility by the nondimensional Froude Number, defined as $Fr = \frac{v^2}{gl}$ where $v$ is the body velocity, $g$ is gravity, and $l$ is the nominal leg length. This was motivated by the \textit{dynamic similarity hypothesis}, which argues that animals move in a dynamically similar fashion when they have speeds proportional to the square root of their leg lengths \cite{alexander1984gaits}. We compile the estimated Froude numbers of quadruped systems contemporary to this work in Table \ref{tab:froude}.

\begin{table}[t!]
    \centering
    \small
    \caption{Reward terms for {\color{black!60!green}{task}}, {\color{blue}stability}, and {\color{orange}smoothness}. Reward from {\small\cite{rudin2021learning}} is adapted to our robot with minor changes.}
    \begin{tabular}{llr}
    \toprule
    Term & Symbol & Equation \\ [0.5ex]
    \midrule
    \tikzmark{task 1 tl}$r_{v^{\footnotesize{\textrm{cmd}}}_x}$: xy velocity tracking  & \( \exp\{{-{|\textbf{v}_{xy}-\textbf{v}^{\text{cmd}}_{xy}|^2} / {\sigma_{vxy}}}\}\)& 0.02 \\ [0.5ex]
     $r_{\omega^{\footnotesize{\textrm{cmd}}}_z}$: yaw velocity tracking & \(  \exp\{{-{(\boldsymbol{\omega}_z-\boldsymbol{\omega}^{\text{cmd}}_{ z})^2} / {\sigma_{\omega z}}}\}\) & 0.01\tikzmark{task 1 br} \\ [0.8ex]
     \tikzmark{task 2 tl}z velocity & \(\textbf{v}_{z}^2\) & -0.04 \\ [0.5ex]
     roll-pitch velocity & \(|\boldsymbol{\omega}_{xy}|^2\) & -0.001 \\ [0.5ex]
     base height & \((h - h^0)^2\) & -0.6 \\ [0.5ex]
     base orientation & \(|\textbf{g}^{\text{ori}}_{xy}|^2\) & -0.002 \\ [0.5ex]
     self-collision & \( \mathbbm{1}_{\text{selfcollision}}\) & -0.02\tikzmark{task 2 br} \\ [0.5ex]
     joint limit violation & \( \mathbbm{1}_{q_i>q_{max} || q_i < q_{min}}\) & -0.2\tikzmark{task 2 br} \\ [0.8ex]
     \tikzmark{task 3 tl}joint torques & \(|\boldsymbol{\tau}|^2\) & -2e-7 \\ [0.5ex]
     joint accelerations & \(|\ddot{\textbf{q}}|^2\) & -5e-9 \\ [0.5ex]
     action rate & \(|\textbf{a}_{t-1} - \at|^2\) & -2e-4 \\ [0.5ex]
     foot airtime & \(\sum t_{air} * \mathbbm{1}_{\text{new contact}}\) & \(0.02\tikzmark{task 3 br}\) \\ [0.5ex]
     \bottomrule
    \end{tabular}
    \label{tbl:rewards}
\end{table}

\DrawBox[thin, draw=black!60!green, dashed]{task 1 tl}{task 1 br}
\DrawBox[thin, draw=blue, dashed]{task 2 tl}{task 2 br}
\DrawBox[thin, draw=orange, dashed]{task 3 tl}{task 3 br}

\end{appendix}

\end{document}